\documentclass[10pt,twocolumn,letterpaper]{article}

\usepackage[algorithms]{wacv}      

\usepackage{graphicx}
\usepackage{amsmath}
\usepackage{amssymb}
\usepackage{booktabs, nicematrix}
\usepackage{colortbl}
\usepackage[accsupp]{axessibility}

\usepackage{times}
\usepackage{epsfig}
\usepackage{tabularx}

\usepackage{multirow}
\usepackage{adjustbox}
\usepackage{float}
\usepackage{xr}

\usepackage{subcaption}
\usepackage{comment}
\usepackage{pifont}

\usepackage[pagebackref=true,breaklinks=true,letterpaper=true,colorlinks,bookmarks=false]{hyperref}

\usepackage[capitalize]{cleveref}
\crefname{section}{Sec.}{Secs.}
\Crefname{section}{Section}{Sections}
\Crefname{table}{Table}{Tables}
\crefname{table}{Tab.}{Tabs.}

\def\wacvPaperID{556} 
\def\confName{WACV}
\def\confYear{2024}
\def\rectimes{\textsc{Recognition}$_{\textsc{Times}}$}
\def\recloc{\textsc{Recognition}$_{\textsc{Location}}$}
\def\recognition{\textsc{Recognition}}
\def\reasoning{\textsc{Reasoning}}
\def\reasontimes{\textsc{Reasoning}$_{\textsc{Times}}$}
\def\reasonloc{\textsc{Reasoning}$_{\textsc{Location}}$}

\begin{document}

\title{Can Vision-Language Models be a Good Guesser? \\
Exploring VLMs for Times and Location Reasoning}

\def\institutey{LMU Munich}
\def\institutew{Munich Center for Machine Learning}
\author{
\textbf{Gengyuan Zhang \textsuperscript{1,2} \quad Yurui Zhang \textsuperscript{3} \quad Kerui Zhang \textsuperscript{1} \quad Volker Tresp \textsuperscript{1,2}}  \\
\textsuperscript{1} LMU Munich, Munich, Germany\\
\textsuperscript{2} Munich Center for Machine Learning, Munich, Germany \\
\textsuperscript{3} Technical University of Munich \\
\tt\small zhang@dbs.ifi.lmu.de 
}
\maketitle

\begin{abstract}
Vision-Language Models (VLMs) are expected to be capable of reasoning with commonsense knowledge as human beings. One example is that humans
can reason where and when an image is taken based on their knowledge. 
This makes us wonder if, based on visual cues, Vision-Language Models that are pre-trained with large-scale
image-text resources can achieve and even surpass human capability in reasoning times and location. 
To address this question, we propose a two-stage \recognition\space \& \reasoning\space probing task applied to discriminative and generative VLMs to uncover whether VLMs can recognize times and location-relevant features and further reason about it.
To facilitate the studies, we introduce WikiTiLo, a well-curated image dataset compromising images with rich socio-cultural cues.
In extensive evaluation experiments, we find that although VLMs can effectively retain times and location-relevant features in visual encoders, they still fail to make perfect reasoning with context-conditioned visual features.
The dataset is available at \url{https://github.com/gengyuanmax/WikiTiLo}.
\end{abstract}


\section{Introduction}

\begin{figure}[htbp]
  \centering
    \includegraphics[width=0.8\linewidth]{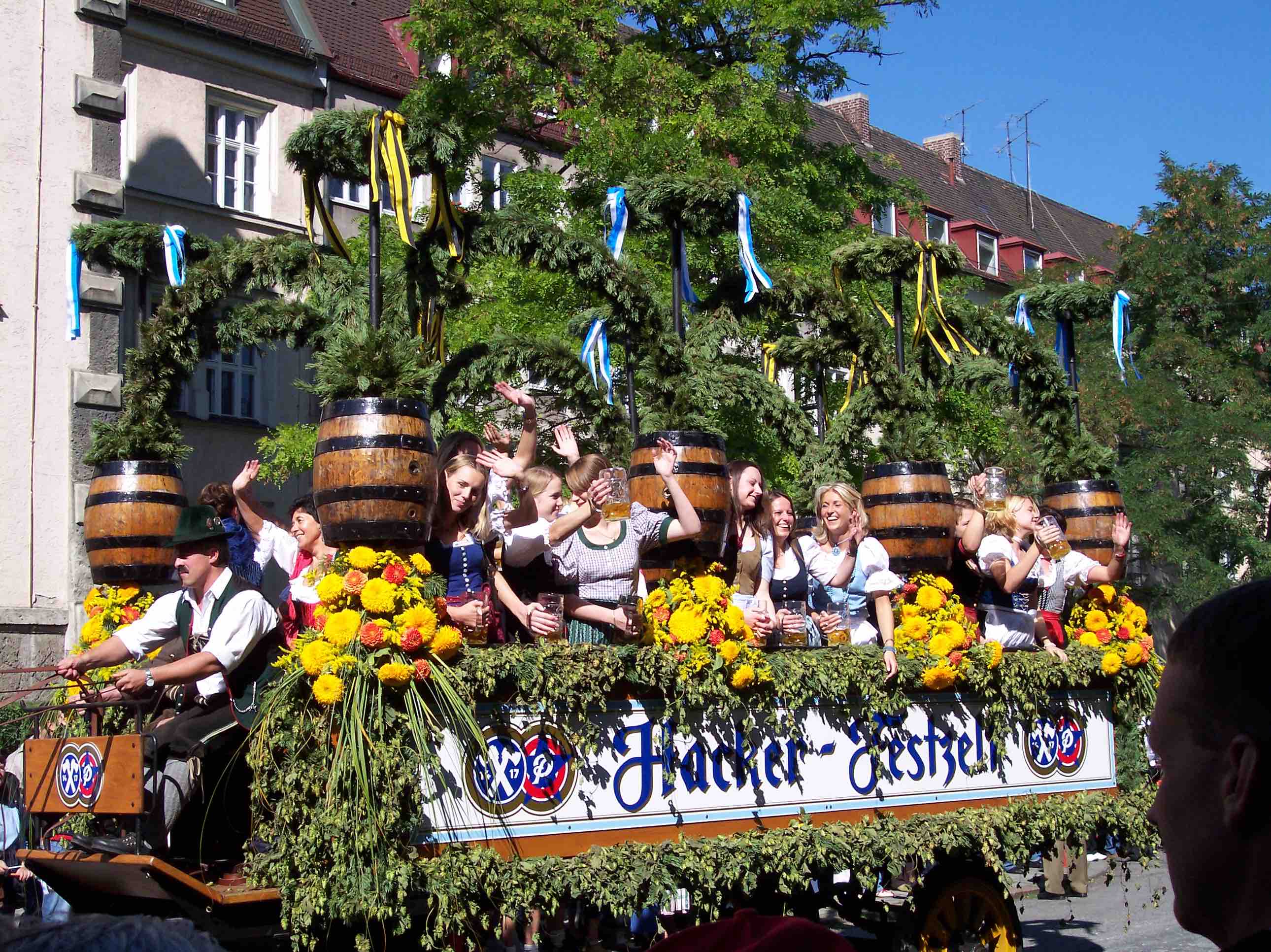}
  \caption{An example image of times and location reasoning in WikiTiLo: Can you tell where and when is this picture taken?
    \label{fig:example}
  }
\end{figure}


Vision-Language Models (VLMs) have exhibited remarkable advancements in enhancing multi-modal learning. On the one hand, discriminative VLMs such as CLIP~\cite{radford2021learning}, BLIP~\cite{li2022blip}, and ImageBind~\cite{girdhar2023imagebind} offer powerful encoders that possess excellent representation and demonstrate significant transferability on recognition and understanding tasks. 
On the other hand, generative VLMs like LLaMA Adapter~\cite{zhang2023llama, zhang2023llamaadapter}, BLIP2~\cite{li2023blip2}, Flamingo~\cite{alayrac2022flamingo}, and LLaVA~\cite{liu2023visual} combine visual encoders from discriminative VLMs with Large Language Models (LLMs) such as LLaMA\cite{touvron2023llama} and GPT3~\cite{brown2020language} to further bridge visual features and leverage the exceptional reasoning abilities of LLMs.

With VLMs trained on extensive knowledge corpora comprising multi-modal data, an intriguing question arises: can VLMs reason about the implicit sociocultural backgrounds associated with an image, such as its times and location?
Just as in the popular game `GeoGuesser,' where players are tasked with locating an image based on visual cues, we contemplate whether VLMs can also exhibit a similar aptitude for being a `good guesser.'
This kind of commonsense understanding necessitates that the model can infer the times and location of an image by comprehending not only the local visual evidence but also the underlying concepts.
For example, from Fig.~\ref{fig:example}, humans with related experience can easily infer that the image was most likely taken in Germany and during the famous Oktoberfest based on the cues such as people wearing traditional Bavarian clothes and celebrating with beers and barrels.



Under this background, our work aims to investigate the times and location reasoning capabilities of VLMs.
Consequently, we put forward two research questions:
\\\textbf{RQ1}: Can discriminative VLMs recognize times and location-relevant features from visual input?\\
\textbf{RQ2}: Can generative VLMs reason about times and locations associated with images based on visual cues?

\begin{figure}
    \centering
    \includegraphics[trim=6cm 6cm 4cm 2cm,clip, width=0.5\textwidth, page=2]{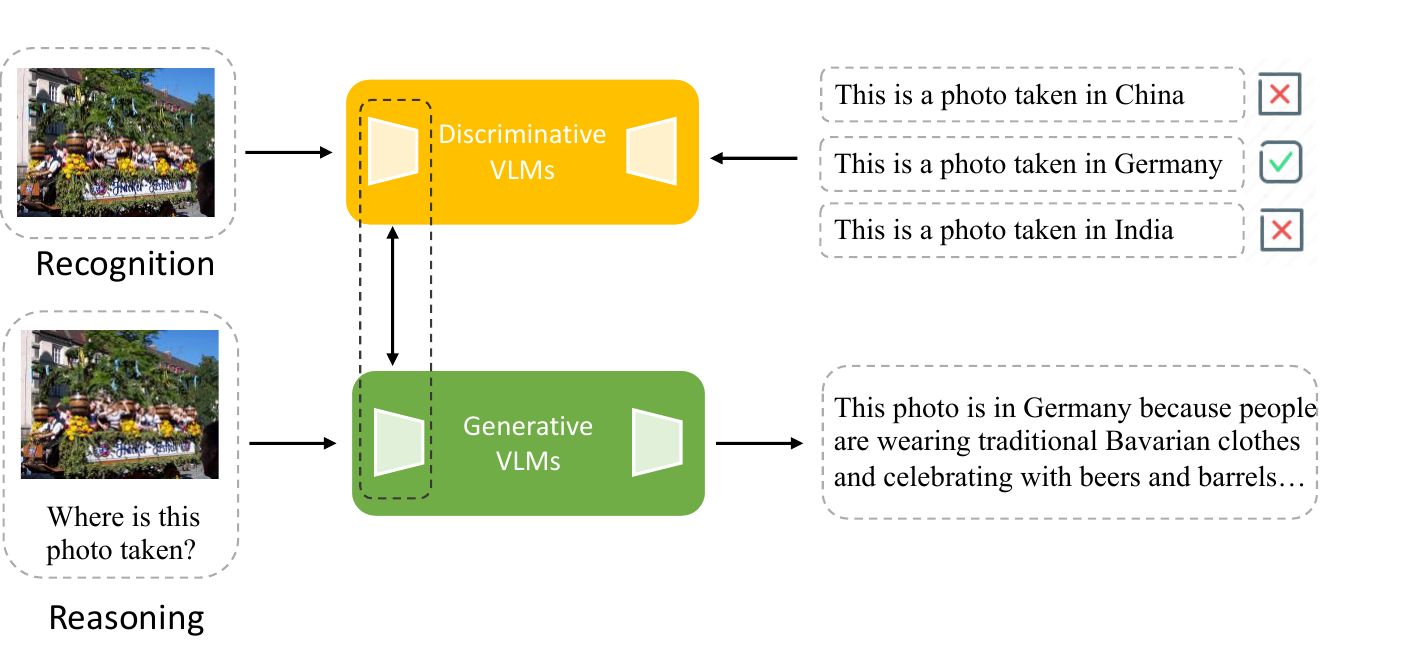}
    \caption{We apply a two-stage probing task \recognition\space and \reasoning\ to both discriminative and generative VLMs. We evaluate \recognition\space on discriminative VLMs and \reasoning on generative VLMs with the visual encoder intact.}
    \label{fig:vlms}
\end{figure}

To address these questions step by step, we devise a two-stage probing task: \recognition\space and \reasoning, applied to discriminative and generative Vision-Language Models (VLMs) as illustrated in Fig.~\ref{fig:vlms}. 
Discriminative and generative VLMs are not just parallel, but generative VLMs are also built upon the encoders of discriminative models.

In the first stage \recognition, we assess whether the visual encoders within discriminative VLMs can successfully identify distinctive features for location and times reasoning using a classification task. Visual features of discriminative VLMs are context-agnostic, which means they are not dependent on tasks and questions.
In the second stage \reasoning, assuming that the visual encoder can identify salient features within its pre-trained representations, we evaluate the ability of generative VLMs, with the visual encoder intact, to conduct times and location reasoning through an open-ended question-answering task. The generative VLMs are now based on context-conditioned visual features and the powerful reasoning ability of large language models.


We also construct a new dataset for times and location reasoning,  WikiTiLo (\textbf{Wiki}Common \textbf{Ti}mes and
\textbf{Lo}cation),
which comprises images captured over a broad time range and is geographically balanced to mitigate cultural bias. 
The dataset has been carefully curated to ensure that each image contains distinct visual cues that align with human expert knowledge.

This is one of the first works to investigate whether state-of-the-art large pre-trained Vision-Language Models, enhanced with Large Language Model techniques, are capable of times and location reasoning.
Our contributions to this work include:
\begin{enumerate}
    \item we construct a new dataset, WikiTiLo, for times and location reasoning, with a focus on the socio-cultural background behind images; 
    \item we propose a two-stage probing strategy, \recognition\space and \reasoning, to evaluate the ability of VLMs to recognize times and location-relevant visual cues and subsequently reason about times and location in a generative setting;
    \item we evaluate three discriminative VLMs and two generative VLMs on this same benchmark. Experiments show that visual encoders in discriminative VLMs can generate context-agnostic visual features that help identify times/locations, but generative VLMs fail to reason based on the visual cues.
\end{enumerate}

\section{Related Work}
\vspace{0.2cm}
\noindent\textbf{Vision-Language Model}
Vision-language models bridge the gap between flourishing studies in computer vision and natural language studies. 
Recent works, including~\cite{chen2020uniter, kim2021vilt, radford2021learning, li2022blip,yang2022unified, lu2019vilbert, li2019visualbert, fu2021violet, yu2022coca, li2022align, yao2021filip, mu2022slip, girdhar2023imagebind} aim to learn a representation of images and texts in a joint feature space and unify vision and language tasks like and visual question answering~\cite{li2022align, kim2021vilt, singh2022flava}.
Along with the boost of pre-training generative Large Language Models (LLM) like GPT-3~\cite{brown2020language}, LLaMA~\cite{touvron2023llama} manifest their excellent commonsense reasoning ability in broad tasks and are fastly adapted to multi-modal scenarios as in works including Flamingo~\cite{alayrac2022flamingo}, LLaMA-adpter~\cite{zhang2023llamaadapter}, BLIP-2~\cite{li2023blip2}.

\vspace{0.2cm}
\noindent\textbf{Times and Location Reasoning}
Recent studies on image times and geo-location prediction~\cite{weyand2016planet, kim2015predicting, seo2018cplanet} show an interesting capacity of deep learning models. 
CLIP~\cite{radford2021learning} also reports a similar Geolocalization classification.
Some geolocation datasets such as Cross-View Time Dataset (CVT)~\cite{salem2020learning}, GT-CrossView~\cite{vo2016localizing}, and Cross-View Image Geolocalization (CVIG)~\cite{lin2013cross} have been provided for the downstream task as Geolocalization. 
But unlike geolocation estimation, our work is more focused on commonsense knowledge-based reasoning and pays more attention to the sociocultural perspective of images when constructing the dataset.
TARA~\cite{fu2022there} is a preceding work on exploring CLIP's commonsense knowledge in this regard; compared to TARA, we construct a dataset consisting of images with a \textbf{wider} timespan and \textbf{unbiased} location distribution and evaluate on a wider range of VLMs.





\vspace{0.2cm}
\noindent\textbf{Model Probing}
Model probing is first proposed in Natural Language Processing to investigate whether representations of Language Models have already learned specific linguistic properties~\cite{shi2016does, hewitt2019designing}.
\cite{basaj2021explaining} proposes a framework to probe visual superpixels as an equivalence of word tokens.
However, probing multimodal models is a relatively new area. \cite{lindstrom2021probing, salin2022are, rosch2023probing, radford2021learning} attempt to investigate multimodal probing tasks such as object counting and position identification to learn whether VLMs are capable of understanding multimodal concepts.
Our work aims to utilize a simple linear probing~\cite{alain2016understanding} method to probe whether VLMs can reason commonsense knowledge like times and location.

\begin{figure*}[h]
  \centering
  \begin{subfigure}[]{0.25\textwidth}
    \centering
    \includegraphics[width=0.8\textwidth,height=2.2cm]{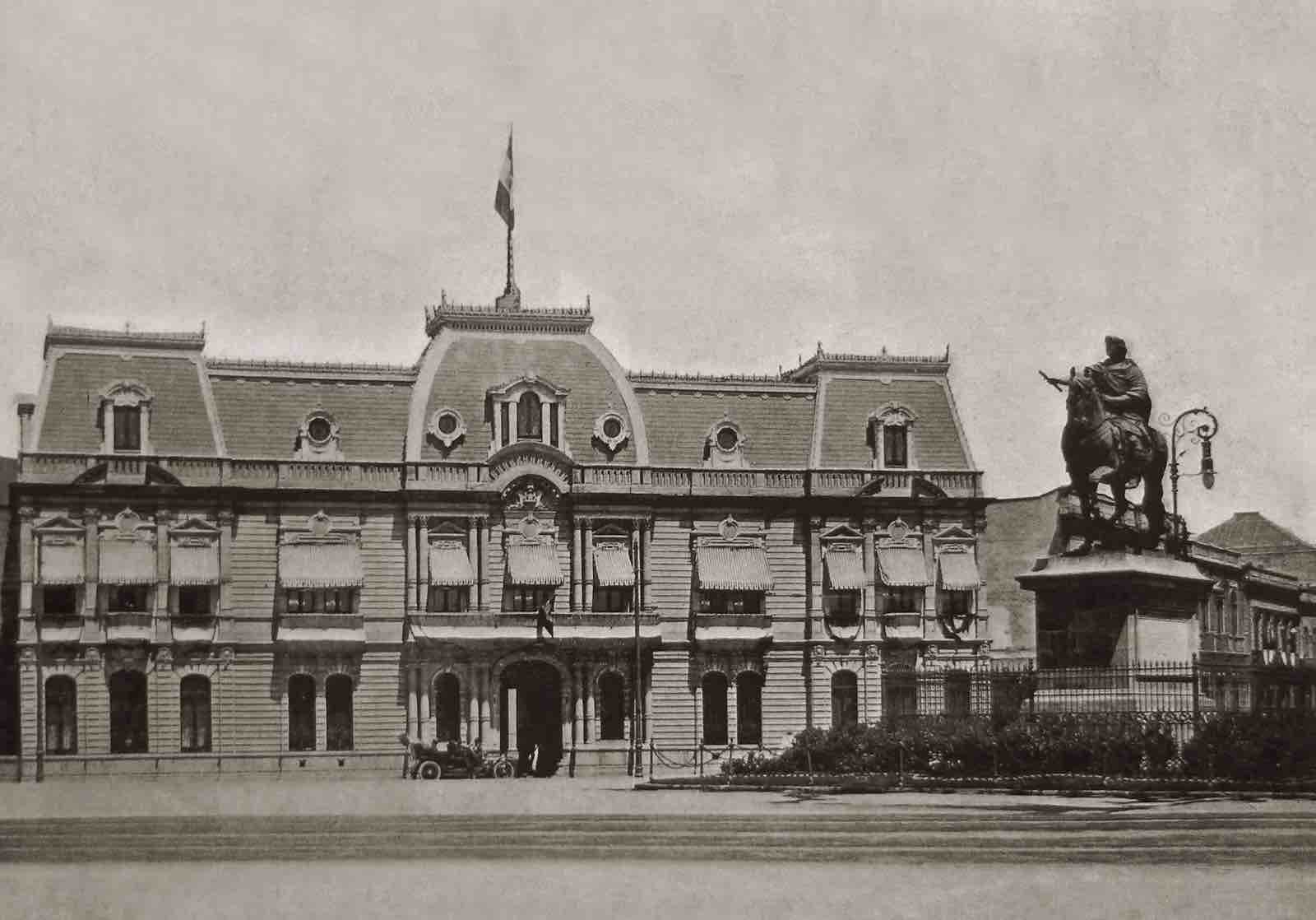}
    \caption{Mexico in 1890}
  \end{subfigure}
  \hfill
  \hspace{-1cm}
  \begin{subfigure}[]{0.25\textwidth}
    \centering
    \includegraphics[width=0.8\textwidth,height=2.2cm]{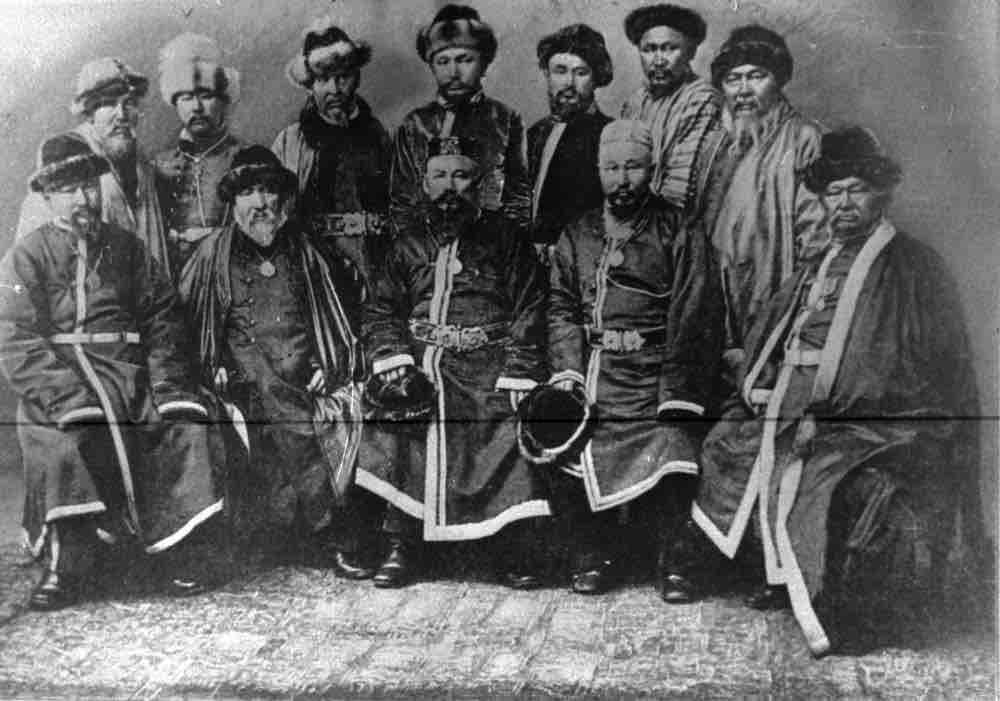}
    \caption{Kazakhstan in 1890}
  \end{subfigure}
  \hfill
  \hspace{-1cm}
  \begin{subfigure}[]{0.25\textwidth}
    \centering
    \includegraphics[width=0.8\textwidth,height=2.2cm]{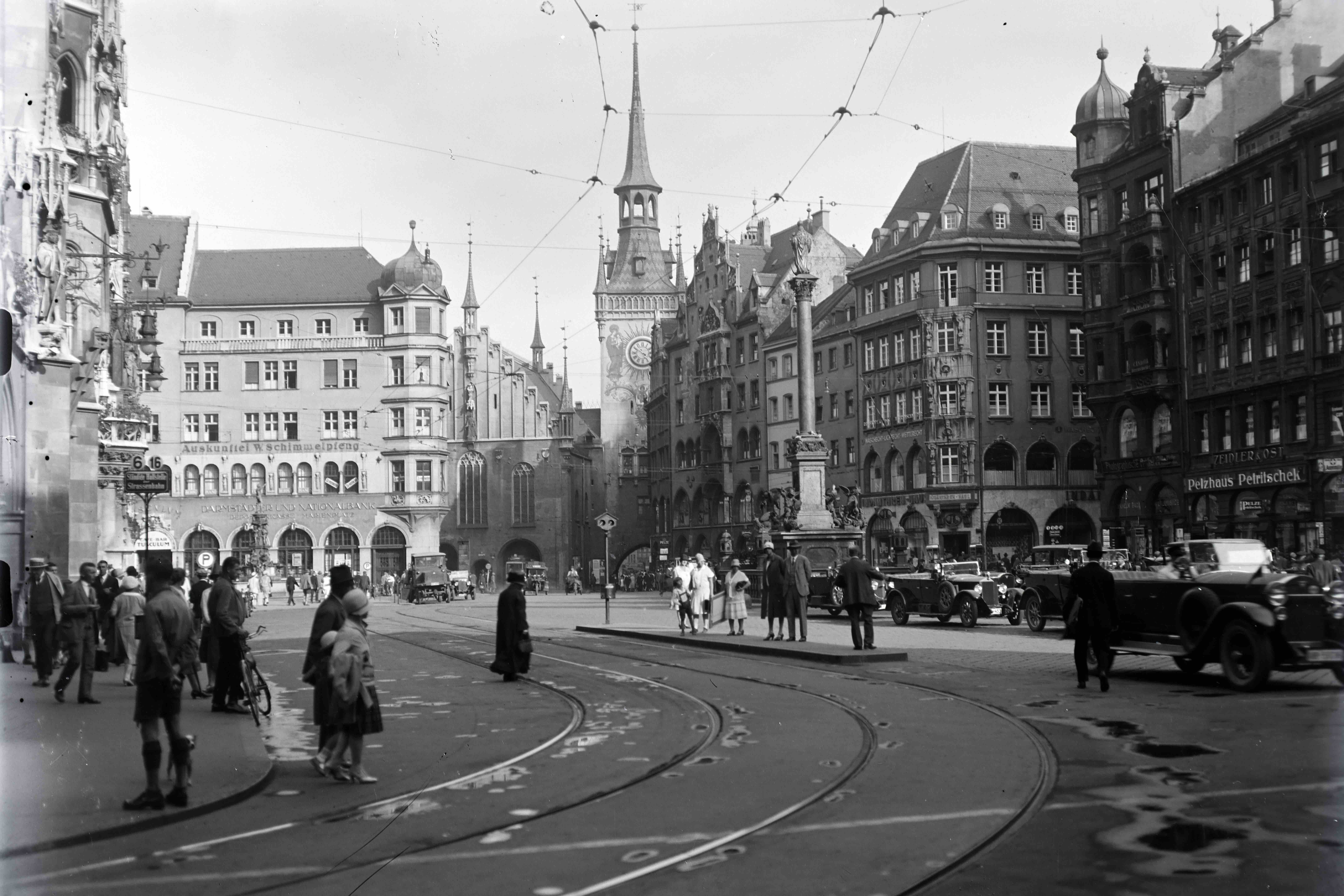}
    \caption{Germany in 1930}
  \end{subfigure}
  \hfill
  \hspace{-1cm}
  \begin{subfigure}[]{0.25\textwidth}
    \centering
    \includegraphics[width=0.8\textwidth,height=2.2cm]{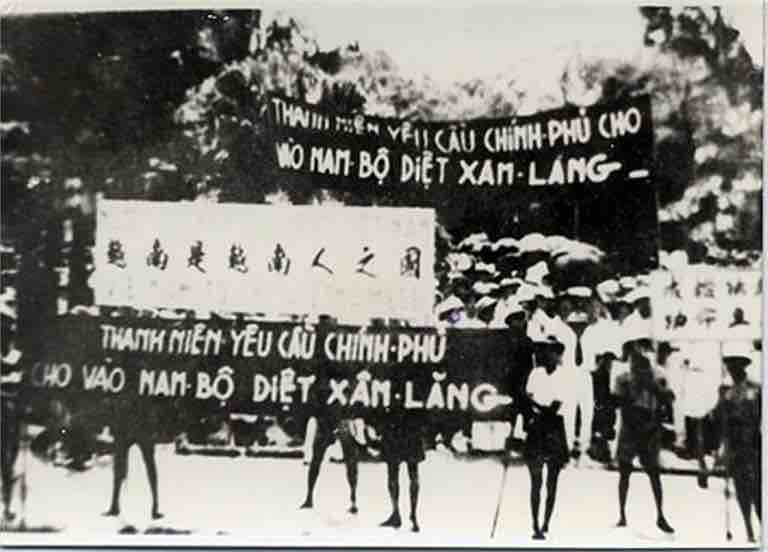}
    \caption{Vietnam in 1940}
  \end{subfigure}

  \begin{subfigure}[]{0.25\textwidth}
    \centering
    \includegraphics[width=0.8\textwidth,height=2.2cm]{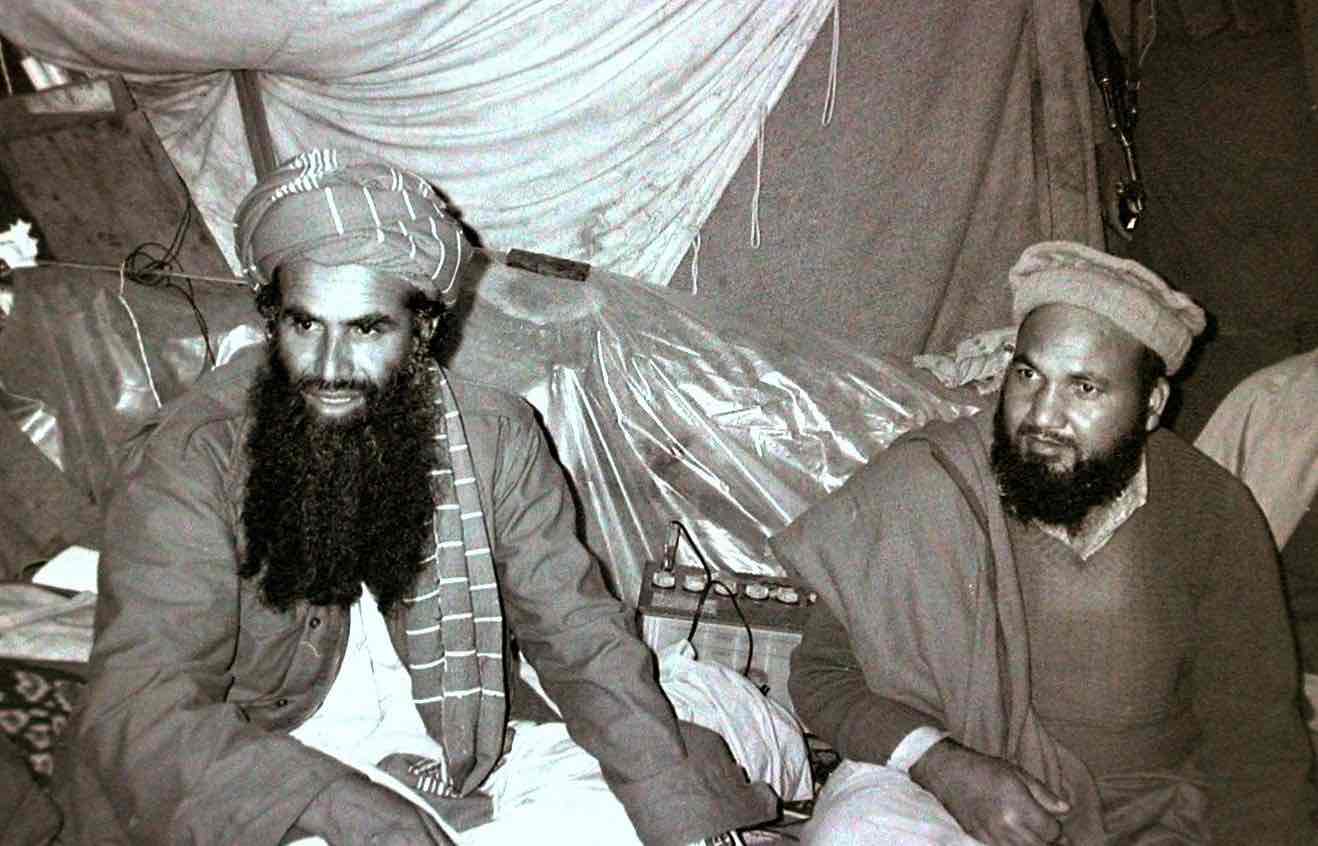}
    \caption{Afghanistan in 1980}
  \end{subfigure}
  \hfill
  \hspace{-1cm}
  \begin{subfigure}[]{0.25\textwidth}
    \centering
    \includegraphics[width=0.8\textwidth,height=2.2cm]{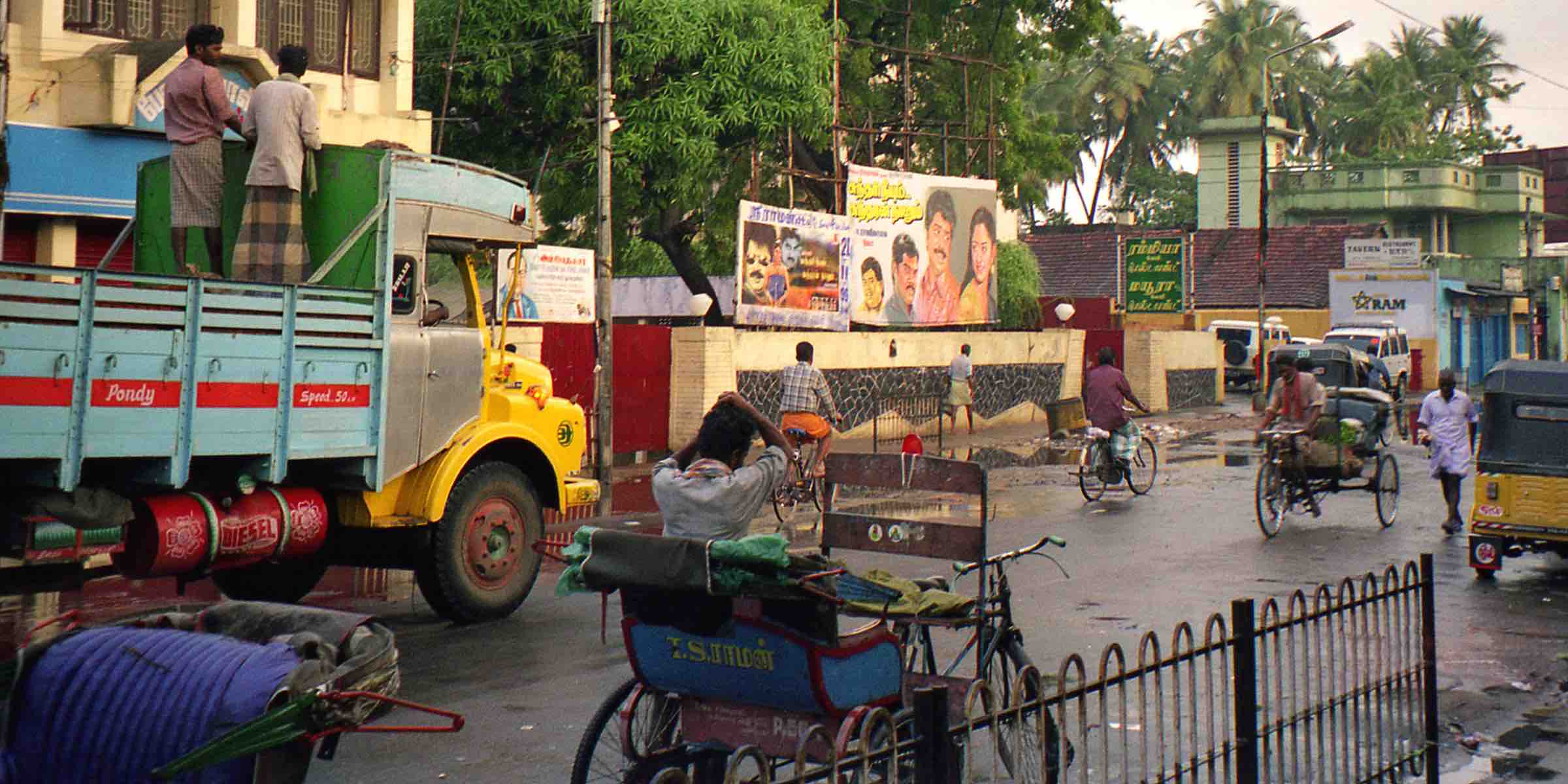}
    \caption{India in 1980}
  \end{subfigure}
  \hfill
  \hspace{-1cm}
  \begin{subfigure}[]{0.25\textwidth}
    \centering
    \includegraphics[width=0.8\textwidth,height=2.2cm]{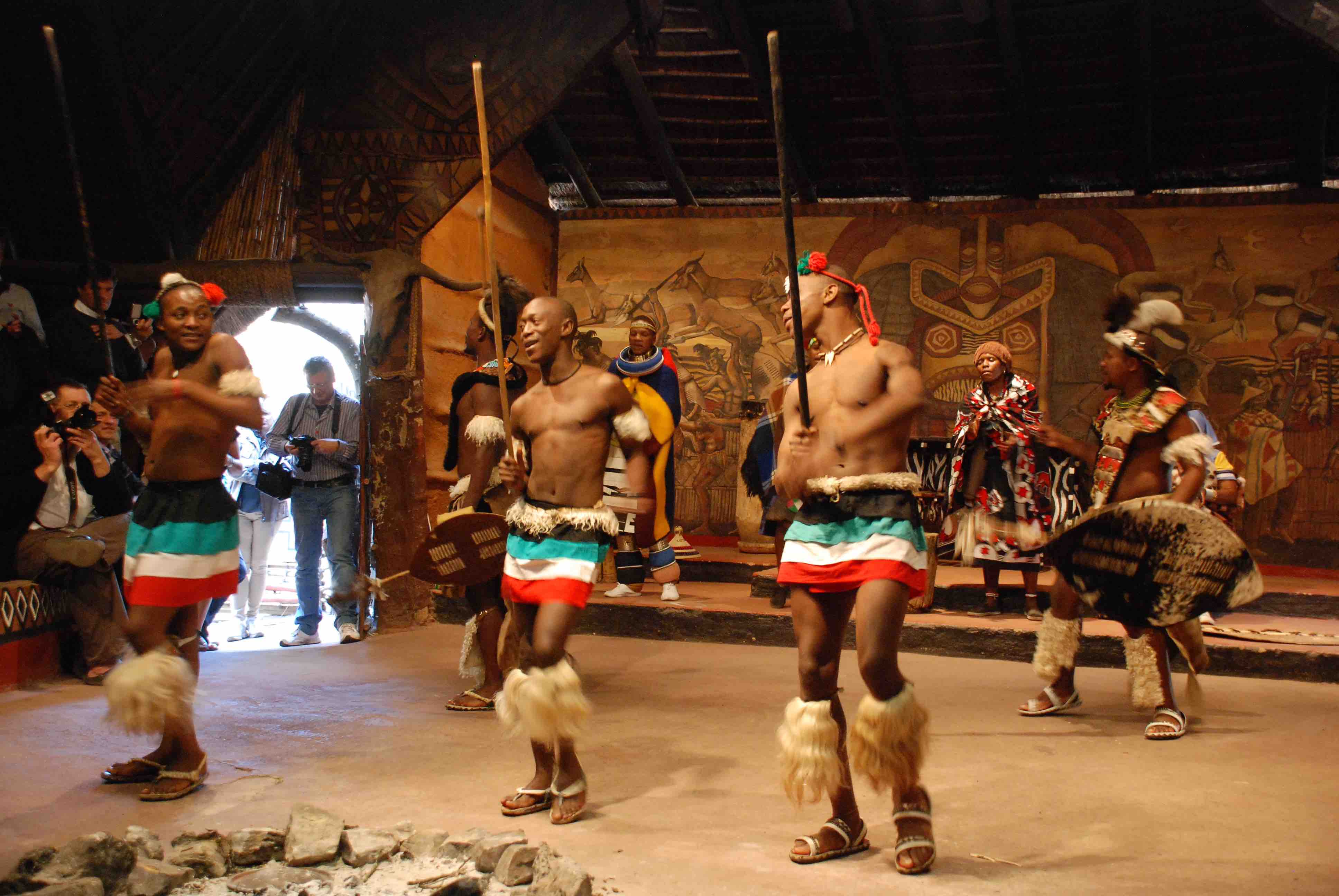}
    \caption{South Africa in 2000}
  \end{subfigure}
  \hfill
  \hspace{-1cm}
  \begin{subfigure}[]{0.25\textwidth}
    \centering
    \includegraphics[width=0.8\textwidth,height=2.2cm]{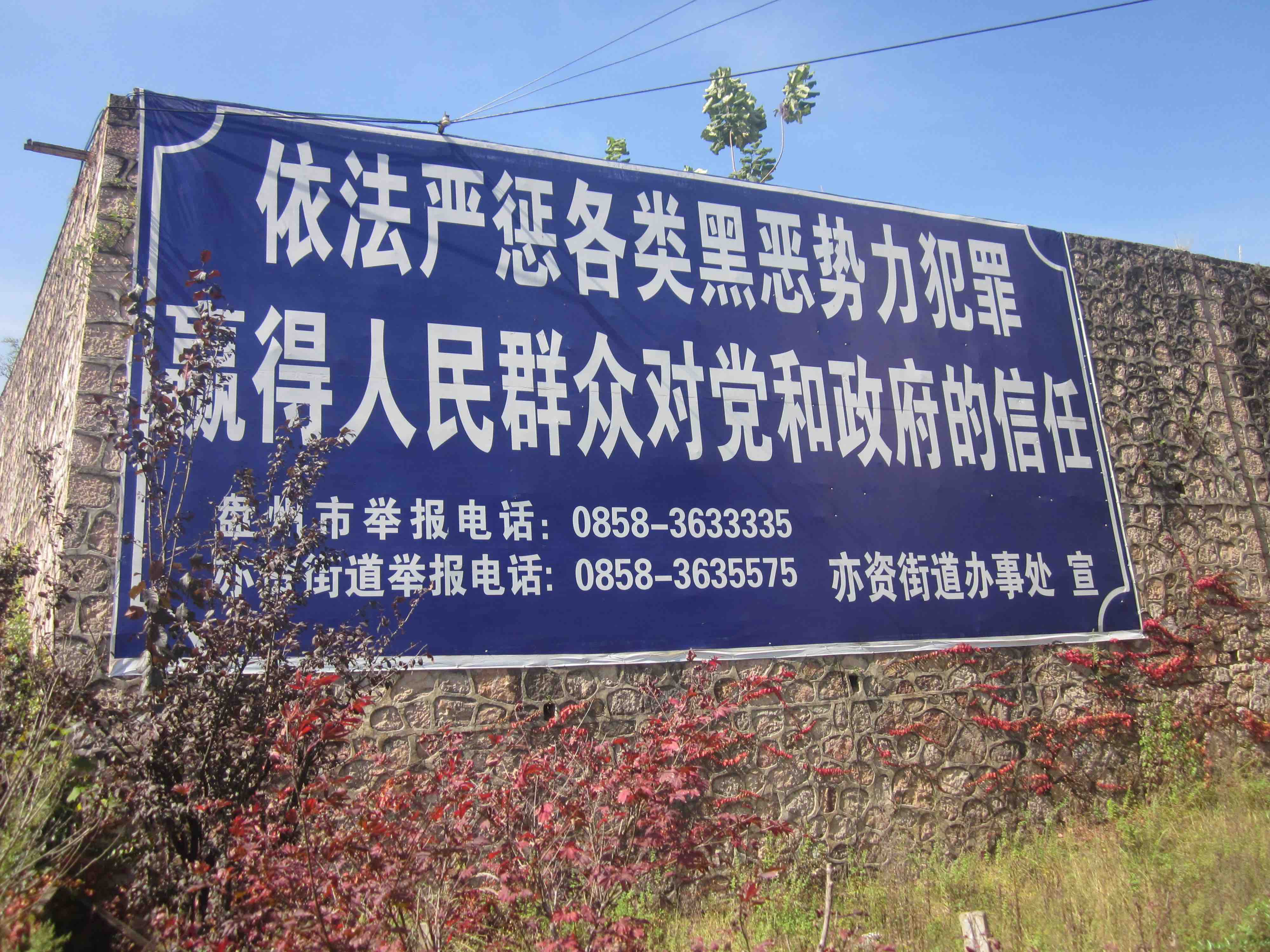}
    \caption{China in 2010}
  \end{subfigure}
  \caption{Some example images in the WikiTiLo exhibit abundant visual cues with a sociocultural background, such as stylish buildings, text on images, and traditional clothing. These visual cues enable humans to reason and draw conclusions based on the provided evidence.}
  \label{fig:9 examples}
\end{figure*}

\section{Dataset: WikiTiLo}
We construct the WikiTiLo (\textbf{Wiki}Common \textbf{Ti}mes and \textbf{Lo}cation) dataset, which consists of 6296 images with annotation of the specific time and country where images are taken. The dataset covers 30 countries in Europe, Asia, America, and Africa and the years from 1826 to 2021. 
We also consider regions as a higher level of geological concept other than countries since the country border does not equate with cultural borders~\cite{taras2016does}. We partition these countries into 8 regions according to UNESCO based on geography and cultural proximity. The designation can be found in Supplementary Material.



\subsection{Data Curation}
Wikimedia Commons\footnote{\tt {https://commons.wikimedia.org/wiki/Main\_Page}} which is a project of the Wikimedia Foundation, contains a media repository of open images, sounds, videos, and other media. The images of Wikimedia Commons are accumulated from different countries and regions around the world, from different historical periods, and in various categories.

We select the images manually based on such a criterion that the location identity and time period features of each
image can be distinguished from architectural patterns, costume styles, languages, social events, photo colors, and quality or other fine-grained features by humans.
In particular, to avoid image distribution bias, we attempt to control image origins and balance the ratio of countries that range from more visible areas like Europe and America to less attended areas like Africa and Central Asia, 
according to socio-cultural characteristics and geographical regions. A detailed selection paradigm is in Supplementary Material.

In total, we have included 30 countries from which images can meet our criteria to represent different regions. 
Fig.~\ref{fig:9 examples} presents some example images in WikiTiLo.

\subsection{Statistics}
The times and location proportions of WikiTiLo are demonstrated in Supplementary Material. All images are nearly evenly distributed in 8 regions. Half of the images are taken after 2000 due to the multimedia era and 10\% are before 1900 due to limited resources and bad quality. For training tasks in the linear probing setting, we use $80\%$ of the entire dataset as the training set, $10\%$ as the validation set, and $10\%$ for the evaluation reported.

\section{Methodology}

\begin{figure}
\vspace{-0.5cm}
    \centering
    \includegraphics[trim=2cm 12cm 12.5cm 1cm,clip, width=0.5\textwidth]{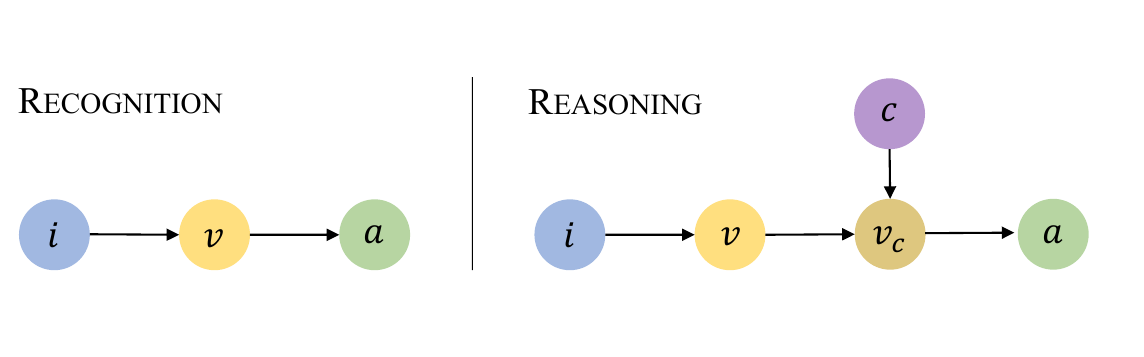}
    \caption{In \recognition, we calculate discriminative VLMs' posterior $\mathbb{P}(a|v)$ of the context-agnostic visual features $v$ of an image $i$. In \reasoning, the context-conditioned visual features $v_c$ are dependent on intact visual features $v$ and question context $c$, including instructions, prompts, and demonstrations; the posterior probability of possible answers is denoted as $\mathbb{P}(a|v_c)$.}
    \label{fig:graph}
\vspace{-0.4cm}
\end{figure}


Given an image $i$, the model estimates the posterior probability of answer space $a$.
As in Fig.~\ref{fig:graph}, in \recognition\space stage, the posterior probability $\mathbb{P}(a)$ is conditioned on context-agnostic visual feature $v$ like visual features of CLIP.
In \reasoning\space stage, $\mathbb{P}(a | v_c)$ is conditioned on context-aware visual feature $v_c$ as in MLLMs, contexts include instructions, interleaved image-text demonstrations, and questions. $v_c$ is conditioned not just on raw images, but also on task-specific and question-related contexts.  

We probe whether context-agnostic visual features of discriminative VLMs contain prominent times/location-relevant cues in the \recognition\space step and probe whether generative VLMs can reason about times and location on context-conditioned visual features in the \reasoning\space.
For both stages, we evaluate models on three tasks, \textbf{Times}, \textbf{Location(Region)}, and \textbf{Location(Country)}, to determine in which era/region/country the photo is taken.


\subsection{\textbf{\recognition}}
In the \recognition\space probing stage, we test whether discriminative VLMs can recognize prominent times/location-relevant visual features via probing the features of visual encoders on multi-class classification tasks. 
\rectimes\ is a fine-grained image classification task where models predict the occurrence time of an image. This task evaluates whether visual representations of a model contain commonsense knowledge of historical times.
\recloc\ is a fine-grained image classification task where models predict the location where an image is taken. We test on both country level and region level that aligns with the dataset information.
We attempt to evaluate the visual encoders of discriminative VLMs in two settings: zero-shot with a prediction calculated as in Eq.~\ref{eq:zs}  and linear probing as in Eq.~\ref{eq:lp}.

\begin{equation}
    \mathbb{P}_{zs}(a_i|v) = \frac{\exp(\mathrm{sim}(\mathrm{VE}(v), \mathrm{TE}(p\circ c_i)))}{\sum_j^{C} \exp(\mathrm{sim}(\mathrm{VE}(v), \mathrm{TE}(p\circ c_j)))}
    \label{eq:zs}
\end{equation}

\begin{equation}
    \mathbb{P}_{lp}(a_i|v) = \mathrm{softmax}(\mathrm{linear}(\mathrm{VE}(v)))_i
    \label{eq:lp}
\end{equation}

\noindent
where $p$ is hard prompts we use, $\mathrm{VE}$ and $\mathrm{TE}$ are visual and textual encoders in VLMs, and $c$ is the class label.

\subsection{\textbf{\reasoning}}\label{sec:reason}
Upon the hypothesis that visual encoders extract recognizable visual features for times and location cues, we further evaluate generative VLMs that can give us good reasoning about times and location based on the cues.
We propose using open-ended visual question-answering tasks \reasontimes\space and \reasonloc\space to reason times and location.

As shown in Fig.~\ref{fig:graph}, the posterior probability of the answer space is conditioned on context-conditioned visual features. A generative VLM generates answers that are conditioned on visual features $V_c$, as in Eq.~\ref{eq:gen}.

\begin{equation}
    \mathbb P_{gen}(a_i|v_c) =  \prod_j \mathbb P(a_{i,j}|\phi(v, c), a_{i,1:j-1})
    \label{eq:gen}
\end{equation}

$\phi$ is a bridge function that projects visual features $v$ and contexts $c$ into the language space, as perceiver resampler in Flamingo~\cite{alayrac2022flamingo}, adapters in LLaMA-adpter~\cite{zhang2023llamaadapter}, and Q-Former in BLIP-2~\cite{li2023blip2}. Generative VLMs can excellently leverage the commonsense reasoning abilities of large language models when predicting the answers.

\begin{figure}[h]
\hspace{-0.1cm}
  \centering
  \scalebox{0.67}{
\fbox{
\centering
\begin{minipage}{35em}
\textbf{OpenFlamingo Cloze Test} \\
\texttt{<image>}
\textit{Output}: This is a local photo taken in country China.
\texttt{<\textbar endofchunk\textbar>}\\
\textbf{Open Flamingo VQA}\\
The photograph was taken in one of the following 30 countries. 30 countries are ... \\
\texttt{<image>}
\textit{Question}: In which country was this photograph taken? \\ 
\textit{Short answer}: China.
\texttt{<\textbar endofchunk\textbar>}\\
\textbf{OpenFlamingo VQA - CoT} \\
The photograph was taken in one of the following 30 countries. 30 countries are ... \\
\texttt{<image>}
\textit{Question}: In which country was this photograph taken? \\
\textit{Answer}: Because in the photo, there is an elderly person wearing a turban standing in a barren land, with houses built on a hill and a snow-capped mountain in the background, which are consistent with the attire and landscape of Afghan, this photo was taken in Afghanistan.\texttt{<\textbar endofchunk\textbar>}
\end{minipage}}}
\caption{We list templates for each protocol for \reasonloc. For OpenFlamingo, few-shot examples are used to facilitate in-context learning. In the case of the zero-shot setting, two text samples are retained but without image tokens \texttt{<image>}.}
\label{fig:prompt}
\vspace{-0.3cm}
\end{figure}

\section{Experiments}

\vspace{0.2cm}
\noindent\textbf{Model selection} For \recognition\space stage, we compare three Vision-Language models: ViLT~\cite{kim2021vilt}, CLIP~\cite{radford2021learning}, and BLIP~\cite{li2022blip}.
In the Linear Probing setting, we also choose one pre-trained ResNet-50~\cite{he2016deep} as a pure vision baseline. 
For \reasoning\ stage, we evaluate two generative VLMs, OpenFlamingo~\cite{anas_awadalla_2023_7733589} and LLaMA-adapter V2~\cite{zhang2023llamaadapter}, both with CLIP-ViT-L/14, which is one of the best-performing encoders in the \recognition\space task.

\vspace{0.2cm}
\noindent\textbf{Evaluation} 
We evaluate models on three tasks as mentioned above: Times, Location(Region), and Location(Country). Answer space for Times includes 4 time periods: pre-1900, 1900-1950, 1950-2000, and post-2000, and includes 8 different regions. For Location(Country), the model should predict in which of the 30 countries the photo is taken.

For all experimental settings, we evaluate the performance using accuracy, precision, and F1 score averaged across classes.
For \recognition, we retrieve the top-1 rank answer as the prediction for all models with the prompts mentioned above.
For \reasoning, considering the free-form responses generated by language models, we post-process all the predictions and filter the keywords that are relevant to the questions. We require an Exact Match for evaluating generative VLMs. 
The only exception is that for \recognition\space on Times, we remap the prediction of LLaMA-Adapter V2-Instruction$^b$ to the time intervals in our settings since strict instruction following is hard to guarantee. Any generated answer that does not directly address the question is considered a mismatch.

\vspace{0.2cm}
\noindent\textbf{Human Performance} serves as an important baseline. Human common sense knowledge about times and location reasoning varies substantially between individuals depending on their background and education. Hence we conducted tests with 12 participants, each assigned 60 randomly sampled images from the test set. All participants have varied backgrounds from six countries across four regions and do not receive any pre-training prior to the test. We report the average accuracy as the human-level baseline.

\begin{figure}
\hspace{-0.1cm}
\centering
\scalebox{0.67}{
\fbox{
\centering
\begin{minipage}{35em}
\textbf{LLaMA-Adapter V2 Instruction$^a$} \\\textit{Instruction}: The photograph was taken in one of the following 30 countries. These 30 countries are ... In which country was this photo taken?\\
\textbf{LLaMA-Adapter V2 Instruction$^b$} \\\textit{Instruction}: In which country was this photo taken? 
\end{minipage}}}
\caption{For the LLaMA-Adapter V2, we use a simple question as the instruction. For the case with labels, the label sentence was added in the instruction before the question. 
}
\vspace{-0.2cm}
\label{fig:instruction1}
\end{figure}

\begin{table*}\tiny
\centering
    \resizebox{0.98\textwidth}{!}{
    \setlength\heavyrulewidth{0.2ex}
    \setlength\lightrulewidth{0.15ex}

    \begin{NiceTabular}{llccccccccc}[colortbl-like]
    \CodeBefore
    \rowcolor{gray!15}{7-12,15-17,20}
    \Body
    \toprule 
    \Block[l]{1-1}{} &
    \Block[c]{1-1}{} &
    \Block[c]{1-3}{\textbf{Times}} & & & 
    \Block[c]{1-3}{\textbf{Country}} & & &
    \Block[c]{1-3}{\textbf{Region}}  
    \\
    \midrule
    \multicolumn{1}{l}{} & \multicolumn{1}{l}{Model} & \multicolumn{1}{c}{Accuracy} & \multicolumn{1}{c}{Precision} & \multicolumn{1}{c}{F1-score} & \multicolumn{1}{c}{Accuracy} & \multicolumn{1}{c}{Precision} & \multicolumn{1}{c}{F1-score} & \multicolumn{1}{c}{Accuracy} & \multicolumn{1}{c}{Precision} & \multicolumn{1}{c}{F1-score} \\
    \midrule 
    \Block[c]{12-1}{\rotatebox[origin=c]{90}{\recognition}} & 
    CLIP-ViT-B/32 & 78.57\% & 70.66\% & 70.66\% & 44.28\% & 43.11\% & 40.19\% & 63.65\% & 67.34\% & 64.42\%\\
    & CLIP-ViT-B/16 & 75.71\% & 67.94\% & 70.08\% & 55.23\% & 54.16\% & 53.71\% & 77.62\% & 77.23\% & 76.86\%\\
    &CLIP-ViT-L/14 & 76.03\% & 69.73\& & 69.62\% & 68.25\% & 60.77\% & 61.87\% & 85.56\% & 86.47\% & 85.72\%\\
    &CLIP-ViT-L/14@336px & \textbf{79.05\%} & \textbf{73.66\%} & \textbf{73.75\%} & \textbf{72.85\%} & \textbf{64.52\%} & \textbf{65.79\%} & \textbf{88.25\%} & \textbf{88.92\%} & \textbf{88.43\%}\\
    &BLIP-129M & 30.95\% & 46.81\% & 46.14\% & 35.23\% & 35.30\% & 30.07\% & 46.51\% & 57.02\% & 49.05\%\\
    &BLIP-129M-Coco & 32.22\% & 50.13\% & 43.49\% & 35.23\% & 34.25\% & 28.74\% & 47.78\% & 52.89\% & 48.14\%\\
    &BLIP-129M-Flickr & 58.57\% & 54.33\% & 56.19\% & 33.49\% & 29.84\% & 27.32\% & 48.89\% & 50.44\% & 47.46\%\\
    &BLIP-ViT-L& 44.44\% & 54.97\% & 52.86\% & \textit{42.22\%} & \textit{43.30\%} & \textit{40.10}\% & 54.60\% & 56.73\% & 54.57\%\\
    &BLIP-ViT-L-Coco & 42.06\% & 51.75\% & 49.40\% & 39.20\% & 38.93\% & 34.61\% & 47.78\% & 52.89\% & 48.14\%\\
    &BLIP-ViT-L-Flickr & \textit{66.51\%} & \textit{58.37\%} & \textit{60.14\%} & 40.47\% & 41.24\% & 37.54\% & \textit{60.00\%} & \textit{61.08\%} & \textit{59.38\%}\\
    &ViLT-Coco & \textit{70.16\%} & \textit{58.51\% }& \textit{57.78\%} & 3.65\% & 3.60\% & 3.10\%  & 16.98\% & 17.39\% & 17.32\%\\
    &ViLT-Flickr30K & 51.90\% & 48.84\% & 49.21\% & \textit{7.93\%} & \textit{4.00\%} & \textit{4.80\%}& \textit{20.32\%} & \textit{20.37\%} & \textit{19.81\%}\\
    \midrule
    \Block[c]{5-1}{\rotatebox[origin=c]{90}{\reasoning}} 
    &OpenFlamingo-Cloze Test & 27.70\% & 26.36\% & 11.49\% & 3.89\% &  3.69\% & 2.18\% & 4.72\% & 8.62\% & 4.72\%\\
    &OpenFlamingo-VQA  & 31.59\% & \textit{30.36\%} & \textit{28.60\%} &  \textit{48.88\%} &  \textit{53.78\%} &  \textit{41.19\%} & 22.49\% & 30.49\% & 18.64\%\\
    &OpenFlamingo-VQA CoT & \textit{35.21\%} & 29.36\% & 28.42\% &  40.3\% &  45.24\% & 33.17\% & \textit{24.04\%} & \textit{39.39\%} & \textit{19.27\%}\\
    &LLaMA-Adapter V2-Instruction$^{a}$ &  \textit{58.02\%} & 28.04\% & 32.88\% & 23.05\% & \textit{52.64\%} & 18.66\% & \textit{19.07\%} & \textit{26.59\%} & \textit{13.01\%}\\
    &LLaMA-Adapter V2-Instruction$^{b}$  & 34.34\% & \textit{58.59\%} & \textit{37.70\%} &  \textit{45.62\%} &  51.57\% &  \textit{35.50\%} & 11.12\% & 10.05\% & 5.99\%\\
    \midrule
    &Frequency baseline &  25.07\% & 25.29\%& 23.27\% & 3.33\% & 2.95\% & 2.88\% & 12.53\% & 12.59\% & 12.24\%\\
    &Human baseline(average) & 67.42\% & - & - & 48.30\% & - & - & 62.42\% & - & - \\
\bottomrule
\end{NiceTabular}}
\caption{Results of performance without training for \textsc{Recognition} and \textsc{Reasoning}. The highest performance among all models is highlighted in \textbf{bold}, and the best performance of each VLM is marked in \textit{italic}. }
\label{tab:zs}
\vspace{-0.1cm}
\end{table*}

\subsection{Experimental Setups}

\vspace{0.2cm}
\noindent\textbf{Zero-shot Setting}
To evaluate the zero-shot performance of discriminative VLMs, we adopt a prompt ``\textit{A historical/recent/contemporary image with black-and-white/color taken in [times period]}" for \rectimes\space classification and ``\textit{a local photograph from [country] in [region]}" for \recloc\space classification. We calculate the image-text similarities across all candidates and retrieve the top rank as predicted classes.


\vspace{0.2cm}
\noindent\textbf{Linear Probing Setting}
We used a single linear layer with vision encoders' embedding size as input size.
The class size is 4 for predicting times, 8 for regions, and 30 for countries.
The visual encoders of the model were frozen during training. Each linear probe was trained for 18 epochs. We search the hyperparameter, including learning rate in the range from $0.05$ to $0.0001$ and learning rate decay among $[0.1, 0.2, 0.5]$ with ray.tune\footnote{\tt{https://www.ray.io/ray-tune}}.
For each task, we use CE loss.

\vspace{0.2cm}
\noindent\textbf{Generation Setting}
We employ two evaluation protocols for OpenFlamingo: VQA and Cloze Test. In both protocols, we utilize either in-context demonstrations or explicit prompts to constrain the generated answer to be within the set of candidate answers. If the generated answer fails to directly address the given question, it is considered incorrect. The results were averaged on 4 trials with different random seeds.
We also conduct tests with varying numbers of in-context shots in the range of $[0, 4, 8, 16, 32]$. 
In the zero-shot setting, we retain two text samples while excluding image tokens \texttt{<image>} and image samples.
To solicit rationale, we also attempt to output Chain-of-Thought(CoT) for prediction by annotating a subset of samples with rationales, which is showcased in Supplementary Material. 
For LLaMA-Adapter V2, we use two different instructions for question answering on times and location reasoning.
Details of all the prompts used can be found in Fig.~\ref{fig:prompt} to Fig.~\ref{fig:instruction1}.

\begin{table}[h]
\centering
    \resizebox{0.4\textwidth}{!}{
    \setlength\heavyrulewidth{0.2ex}
\setlength\lightrulewidth{0.15ex}

\begin{NiceTabular}{lccc}[colortbl-like]
    \CodeBefore
    \rowcolor{gray!15}{3-9}
    \rowcolor{gray!15}{16-17}
    \Body
    \toprule 
Model & Accuracy & Precision & F1-score \\
\midrule 
ResNet-50 & 80.63\% & 73.71\% & 71.89\% \\
CLIP-RN50 & 85.87\% & 81.26\% &	80.73\%  \\
CLIP-RN101 & 86.67\% & 80.71\% & 80.71\% \\
CLIP-RN50x16 & 90.32\% & 86.24\% & 86.39\% \\ 
CLIP-ViT-B/32 & 89.37\% & 85.67\% & 85.04\%\\
CLIP-ViT-B/16 & 88.25\% & 83.04\% & 82.95\%\\
CLIP-ViT-L/14 & 90.63\% & 86.20\% & 86.14\%\\
CLIP-ViT-L/14@336px & \textbf{92.70}\% & \textbf{89.40}\% & \textbf{89.64}\%\\
BLIP-129M & 86.51\% & 81.36\% & 80.51\%\\
BLIP-129M-Coco & 85.87\% & 81.16\% & 80.77\%\\
BLIP-129M-Flickr & 86.67\% & 82.42\% & 80.91\%\\
BLIP-ViT-L & \textit{88.41\%} & \textit{84.72\%} & \textit{84.02\%}\\
BLIP-ViT-L-Coco & 86.67\% & 82.45\% & 81.47\%\\
BLIP-ViT-L-Flickr & 87.94\% & 83.73\% & 83.23\%\\ 
ViLT-Coco & 80.00\% & 75.05\% & 72.72\%\\ 
ViLT-Flickr30K & \textit{80.79\%} & \textit{75.86\%} & \textit{73.19\%}\\ 
\bottomrule
\end{NiceTabular}}
\caption{Results of Linear Probing for Times}
\label{tab:lp_times}
\end{table}

\section{Results}
\subsection{\textbf{\recognition}}
We compare the model performance on zero-shot and linear probing in \recognition \ stage.

\vspace{0.2cm}
\noindent\textbf{Zero-shot Performance} 
Tab.~\ref{tab:zs} shows the results of the zero-shot performance of VLMs on \rectimes\space and \recloc\space  tasks.
We notice that CLIP variants achieve impressive performance on both tasks and outperform other models by a large margin, which corresponds with the findings of excellent zero-shot performance in \cite{radford2021learning}. Amongst all variants, model performance on times classification does not vary much from model architecture.
By contrast, BLIP performs poorly on both tasks.

We also find that among each VLM family, models with a larger visual encoder perform slightly better than others on times classification but have a bigger advantage on location classification. 
Datasets also are critical for model performance. CLIP is trained on large-scale image-text data and manifests the best capabilities of reasoning times and location. However, we still cannot ground the failure of ViLT and BLIP fully since the domain shift between pre-training datasets and WikiTiLo might be huge.  

Compared to human performance, all CLIP models outperform our testee on both tasks. 
Noteworthily, human performance varies individually according to their background, with a variance of $10.41\%$ in times classification and $8.69\%$ in location classification.

\vspace{0.2cm}
\noindent\textbf{Linear Probing Performance} 
Tab.~\ref{tab:lp_times}-\ref{tab:lp_location} shows the results of model performance on the Times and Location(Region) task in the linear probing setting. We report the models' best performance after the hyperparameter search.
We use pre-trained ResNet-50 as a pure-vision baseline. ResNet-50 shows non-inferior performance on the Times task but cannot compare to most Vision-Language models on the Location(Region) except ViLT. 

\begin{table}[h]
\centering
    \resizebox{0.4\textwidth}{!}{
    \setlength\heavyrulewidth{0.2ex}
\setlength\lightrulewidth{0.15ex}

    \begin{NiceTabular}{lccc}[colortbl-like]
    \CodeBefore
    \rowcolor{gray!15}{3-9}
    \rowcolor{gray!15}{16-17}
    \Body
    \toprule 
Model & Accuracy & Precision & F1-score \\
\midrule 
ResNet-50 & 54.13\% & 54.85\% & 54.45\% \\
CLIP-RN50  &  72.06\% & 71.98\% & 72.04\% \\
CLIP-RN101  & 74.44\% &	74.44\% & 74.41\%\\
CLIP-RN50x16  & 86.67\% & 87.31\% & 86.93\% \\ 
CLIP-ViT-B/32 & 79.37\% & 79.36\% & 79.11\%\\
CLIP-ViT-B/16 & 84.60\% & 85.18\% & 84.88\%\\
CLIP-ViT-L/14 & 90.95\% & 91.36\% & 90.92\%\\
CLIP-ViT-L/14@336px & \textbf{93.33}\% &\textbf{ 93.90}\% & \textbf{93.37}\%\\
BLIP-129M & 75.08\% & 75.10\% & 75.26\%\\
BLIP-129M-Coco & 77.14\% & 76.77\% & 76.99\%\\
BLIP-129M-Flickr & 76.98\% & 77.01\% & 76.81\%\\
BLIP-ViT-L & 76.51\% & 77.37\% & 76.77 \%\\
BLIP-ViT-L-Coco & 78.10\% & 77.92\% & 78.04\%\\
BLIP-ViT-L-Flickr & \textit{78.73\%} & \textit{78.64\%} & \textit{78.53\%}\\ 
ViLT-Coco & 42.70\% & \textit{45.61\%} & 42.12\%\\ 
ViLT-Flickr30K & \textit{45.08\%} & 42.91\% & \textit{43.01\%} \\
\bottomrule
\end{NiceTabular}}
\caption{Results of Linear Probing for Location(Region)}
\label{tab:lp_location}
\end{table}

All models enjoy a prominent performance improvement compared to their zero-shot baselines.  
CLIP models still achieve much better performance than other VLMs. 
In particular, BLIP models also enjoy a large improvement in both tasks. 
Thus, we deduce that both CLIP and BLIP have learned comparably informative and discriminative features for these two tasks in pre-training already, which profits from their pre-training data scale.

We conclude that reasoning location is a harder and more interesting task than reasoning times. This could be explained by the fact that location classification is a more fine-grained task, which requires the model to distinguish more detailed visual cues at a commonsense level, like understanding distinct geological and cultural elements.


We also find that visual encoders have an impact on the linear probing results. ViLT still shows a poor performance and is even worse than ResNet-50. Since CLIP-RN50 with the same encoder structure and BLIP trained with the same datasets still shows a good performance, we can only attribute it to ViLT with a simple linear patch embedding cannot produce informative visual features for such tasks. 

\begin{figure*}[h]
\vspace{-1.3cm}
\begin{subfigure}[b]{0.24\textwidth}
\includegraphics[width=\textwidth, height=3cm]
{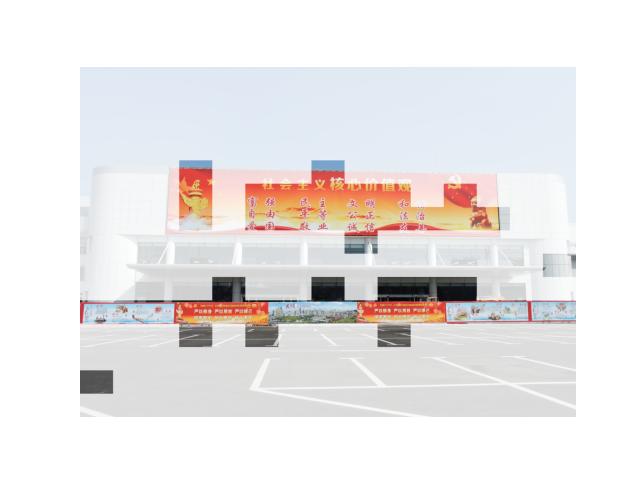}
\vspace{-0.8cm}
\caption{}
\end{subfigure}
\begin{subfigure}[b]{0.24\textwidth}
\includegraphics[width=\textwidth, height=3cm]{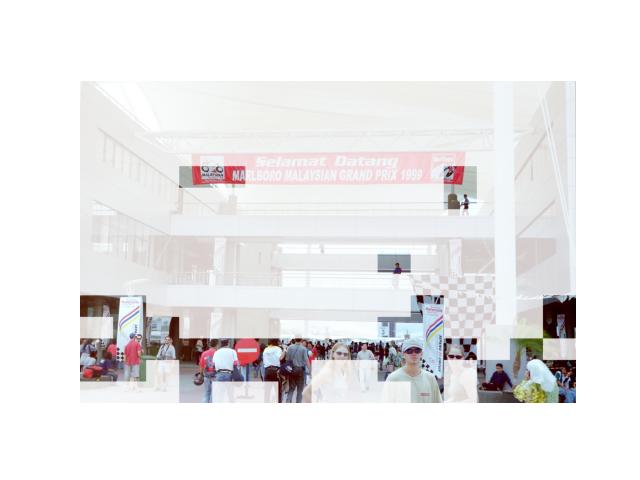}
\vspace{-0.8cm}
\caption{}
\label{vilt:malaysia1990}
\end{subfigure}
\begin{subfigure}[b]{0.24\textwidth}
\includegraphics[width=\textwidth, height=3cm]{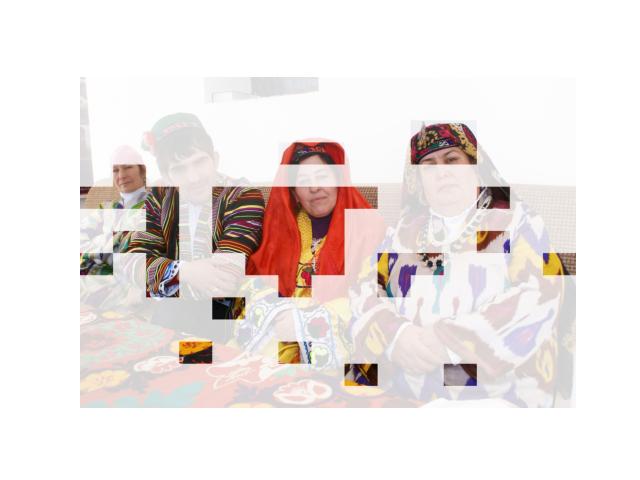}
\vspace{-0.8cm}
\caption{}
\end{subfigure}
\begin{subfigure}[b]{0.24\textwidth}
\vspace{-0.1cm}
\includegraphics[width=\textwidth, height=3cm]{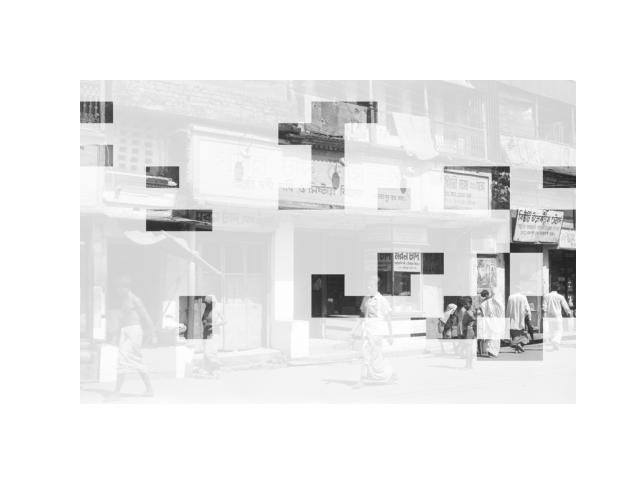}
\vspace{-0.8cm}
\caption{}
\end{subfigure}

\begin{subfigure}[b]{0.24\textwidth}
\vspace{-0.1cm}
\includegraphics[width=\textwidth, height=3cm]{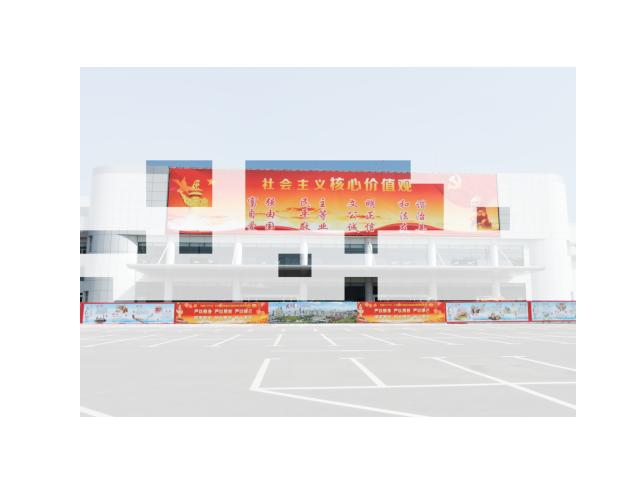}
\vspace{-0.8cm}
\caption{}
\end{subfigure}
\begin{subfigure}[b]{0.24\textwidth}
\vspace{-0.1cm}
\includegraphics[width=\textwidth, height=3cm]{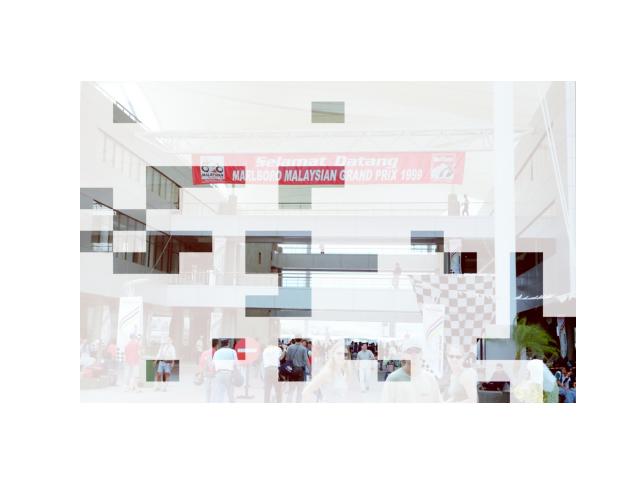}
\vspace{-0.8cm}
\caption{}
\label{clip:Malaysia1990}
\end{subfigure}
\begin{subfigure}[b]{0.24\textwidth}
\includegraphics[width=\textwidth, height=3cm]{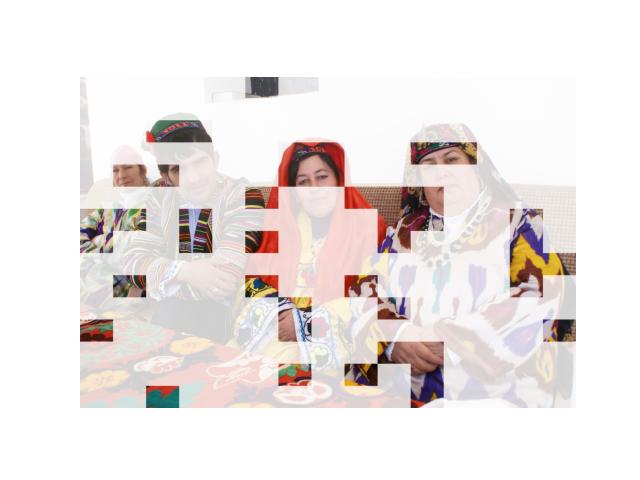}
\vspace{-0.8cm}
\caption{}
\end{subfigure}
\begin{subfigure}[b]{0.24\textwidth}
\includegraphics[width=\textwidth, height=3cm]{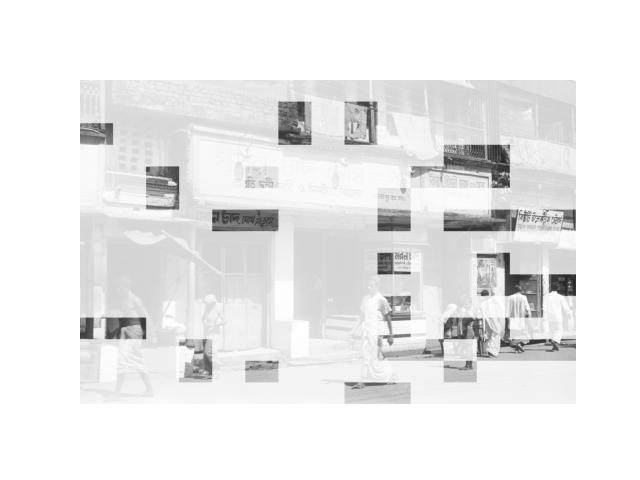}
\vspace{-0.8cm}
\caption{}
\end{subfigure}

\begin{subfigure}[b]{0.24\textwidth}
\includegraphics[width=\textwidth, height=3cm]
{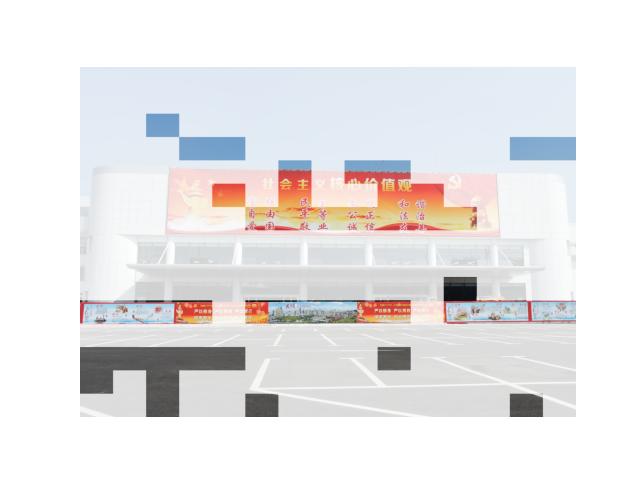}
\vspace{-0.8cm}
\caption{}
\label{clip:china2010}
\end{subfigure}
\begin{subfigure}[b]{0.24\textwidth}
\includegraphics[width=\textwidth, height=3cm]{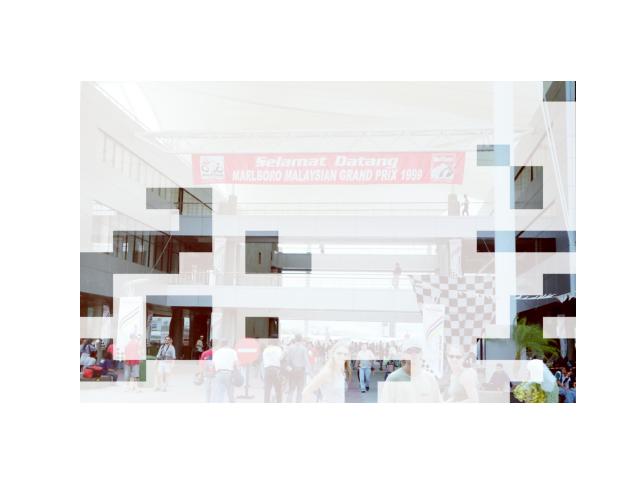}
\vspace{-0.8cm}
\caption{}
\end{subfigure}
\begin{subfigure}[b]{0.24\textwidth}
\includegraphics[width=\textwidth, height=3cm]{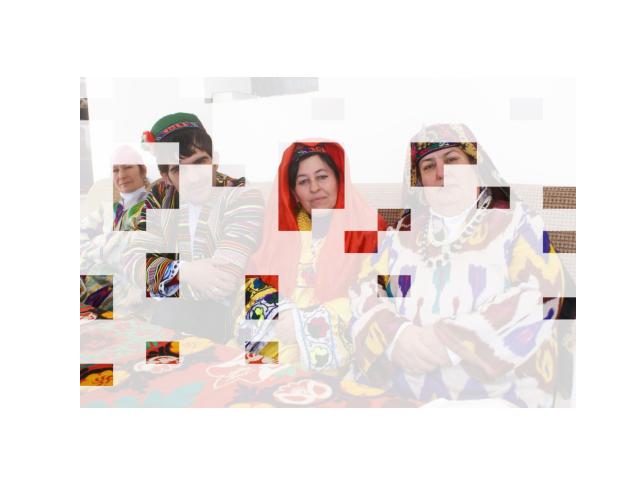}
\vspace{-0.8cm}
\caption{}
\end{subfigure}
\begin{subfigure}[b]{0.24\textwidth}
\includegraphics[width=\textwidth, height=3cm]
{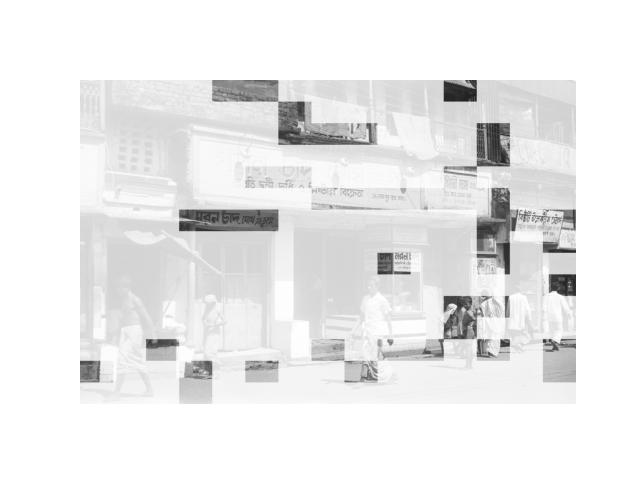}
\vspace{-0.8cm}
\caption{}
\end{subfigure}
\caption{Visualization of transportation plan of word patch alignment on location classification. Best viewed zoomed in. Rows from top to bottom: ViLT, CLIP, and BLIP. Columns from left to right: China (Eastern Asia) in 2010, Malaysia (South-Eastern Asia) in 1990, Tajikistan (Central Asia) in 2000, Bangladesh (Southern Asia) in 1970.}
\label{fig:vis_space}
\end{figure*}

\vspace{0.2cm}
\noindent
\textbf{Visualization}
Inspired by \cite{kim2021vilt}, we adopt a cross-modal alignment of times/location tokens and image patches on linear probing models for visualization, to show that visual encoders can recognize relevant discriminative features. We use the prompts in the zero-shot setting. We select the first $25\%$ of visual patches that have more mass transported from word tokens of labels in prompts to visual patches to emphasize. 
Fig.~\ref{fig:vis_space} and Supplementary Material. show the visualization of location classification.
For Times-relevant questions, the attended patches seem less specific. Generally, visual tokens in the background instead of foreground objects have seemingly dominant contributions. 

Interestingly, we find that for \recloc, models tend to attend to specific regions like scene texts and human clothing. CLIP can locate the Chinese characters on the board, as shown in Fig.~\ref{fig:vis_space}\subref{clip:china2010} and banner texts reading ``Malaysian" in Fig.~\ref{fig:vis_space}\subref{clip:Malaysia1990}. When the model performs poorly, as shown in Fig.~\ref{fig:vis_space}\subref{vilt:malaysia1990}, attended patches are less meaningful. This shows us that VLMs can recognize distinct and discriminative visual cues on an image that help reason commonsense knowledge about the location.

\subsection{\textbf{\reasoning}}




\noindent\textbf{Performance of In-Context Learning-based Reasoning}
From Tab.~\ref{tab:zs}, we also find that OpenFlamingo Cloze Test only achieves frequency baseline results. This reveals that in-context demos can teach the generative LLMs the mapping to output label space but do not leverage any semantic information from visual features.
In the VQA protocol, OpenFlamingo has higher accuracy compared to the Cloze Test protocol.
However, enforcing the model to output Chain-of-Thought does not improve the performance. 
Regarding the impact of the number of shots in in-context learning, we show that more shots do not improve performance on location reasoning substantially, as in Supplementary Material. Especially for \reasontimes, we find the output prediction is more unstable when having more in-context shots and deteriorates the performance.
\textit{Interestingly}, we find that models perform even worse on reasoning regions than countries and the language models fail to relate the countries to their affiliated regions in some cases, of which we include a case study in Supplementary Material.

\vspace{0.2cm}
\noindent\textbf{Performance of Instruction-based Reasoning}
LLaMA-Adapter V2 exhibits unsatisfying performance compared to other models in general.
With Instruction$^a$, the model's performance declines when provided with label instructions.
With Instruction$^b$, the model achieves performance that is comparable to OpenFlamingo VQA as well as the averaged human baselines. We also find different instructions have a big impact on the model performance on \reasoning.

\subsection{Analysis}
\textbf{We notice the performance gap between \recognition\space and \reasoning}. Generative models also can hardly achieve averaged human baselines, which cannot live up to the expectations since large-scale VLMs should benefit more from a massive knowledge base in our assumption.

Based on the fact that in \recognition\space stage, good performance of discriminative VLMs shows visual encoders can extract good visual features even without any task contexts, we reasonably speculate that the performance gap lies in two aspects. 
(1) Context-conditioned visual features $v_c$ cannot retain answer-relevant information, and (2) large language models fail to reason based on the visual cues. 

Since for most generative VLMs, the projection modules are deeply coupled with the language models, like adapters used in LLaMA-adapter, it is hard to probe the context-conditioned visual features $v_c$ via linear prober. We conduct a qualitative study on failure cases.
To validate our hypotheses, we conducted a case study on the failure cases.

\begin{figure}[h]
\vspace{-0.3cm}
    \hspace{-0.6cm}
      \includegraphics[trim=0 0 5cm 0, clip, width=0.54\textwidth]{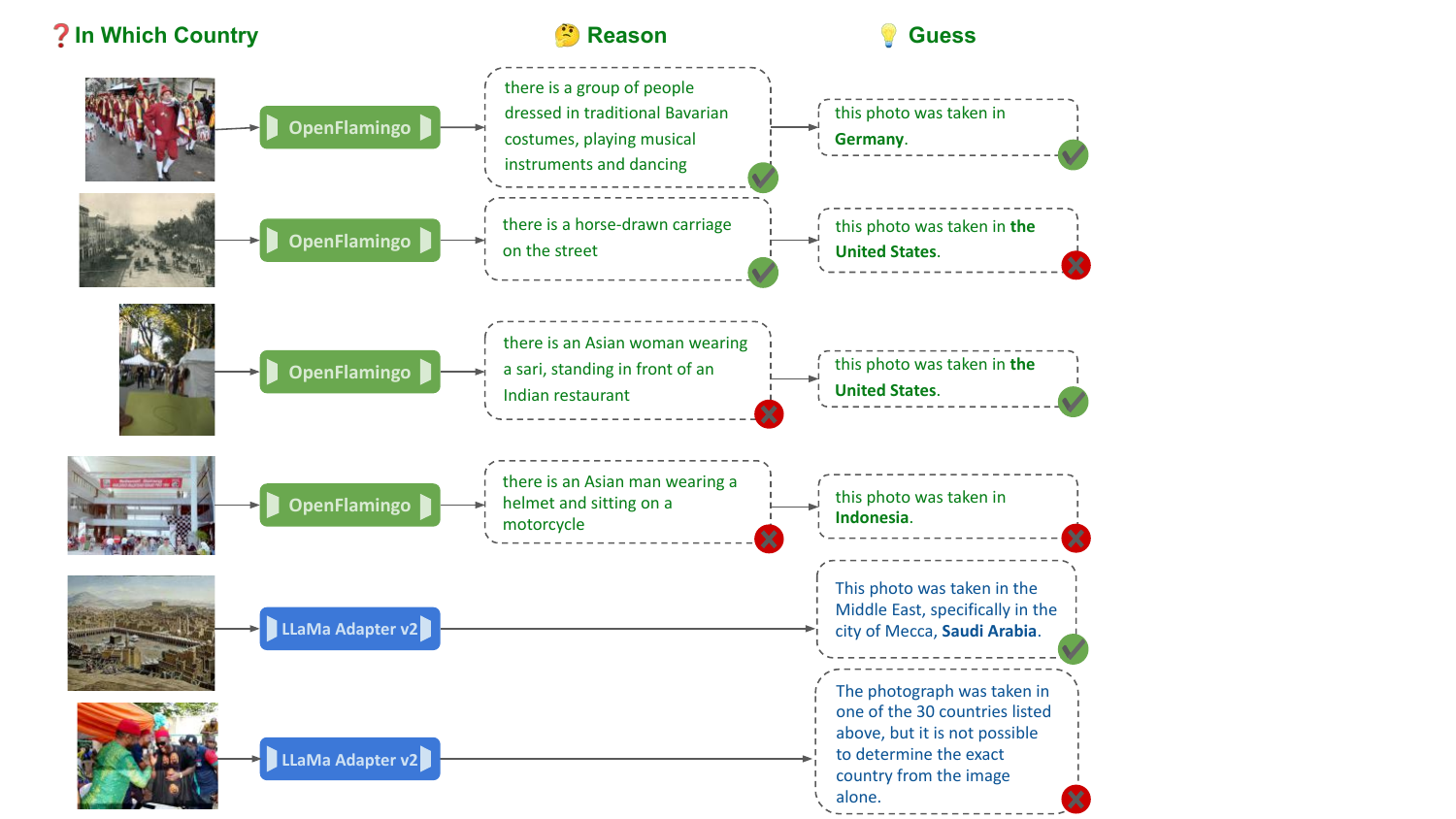}
    \caption{Example cases of interest from OpenFlamingo VQA with Chain of Thought, and LLaMA Adapter V2. We find that the answers generated are not grounded on the visual cues and the reasoning and predictions are not coherent, which both fail.}
    \label{fig:case-study}
  \end{figure}

\vspace{0.2cm}
\noindent
\textbf{Failure Cases}
We investigate the failure cases in the \reasoning\ phase of generative VLMs. In Fig.~\ref{fig:case-study}, we analyze different types of failure cases for OpenFlamingo and LLaMA-Adapter V2. 
By requiring OpenFlamingo to output Chain-of-Thought and providing clear instructions to LLaMA-Adapter V2, both models can make grounded guesses.
However, failure guesses can occur in the following scenarios: 1) when OpenFlamingo fails to generate a grounded reason based on an image, and 2) when OpenFlamingo cannot generate coherent answers based on the reason. 
We find that these failures are primarily caused by OpenFlamingo being heavily influenced by the few-shot demos, which limits its ability to extract relevant and useful information to a significant extent. On the other hand, for LLaMA-Adapter V2, predictions are rejected when the model fails to locate relevant visual cues in an image, despite humans being able to reason effectively. 
This highlights that generative VLMs still struggle to fully leverage the visual cues in images for times and location reasoning. 


\vspace{0.2cm}
\noindent
\textbf{Dataset Bias}
Humans sometimes rely on extraneous information such as image quality, color, or style to determine times of photo-taking. Considering the images in WikiTiLo come from different media resources, it is crucial to investigate whether the performance of VLMs will also be affected by these factors. We compare 8 models, including 3 discriminative VLMs and 5 generative VLMs, on the original test set, and 3 style-transferred test sets: images in lower quality, in gray scale, and in sketch-version. We assume that if the reasoning is based on image style bias instead of image details, the model results will deteriorate after style transfer.

\begin{figure}
\vspace{-0.3cm}
\centering
\begin{subfigure}{0.48\textwidth}
    \includegraphics[width=\textwidth]{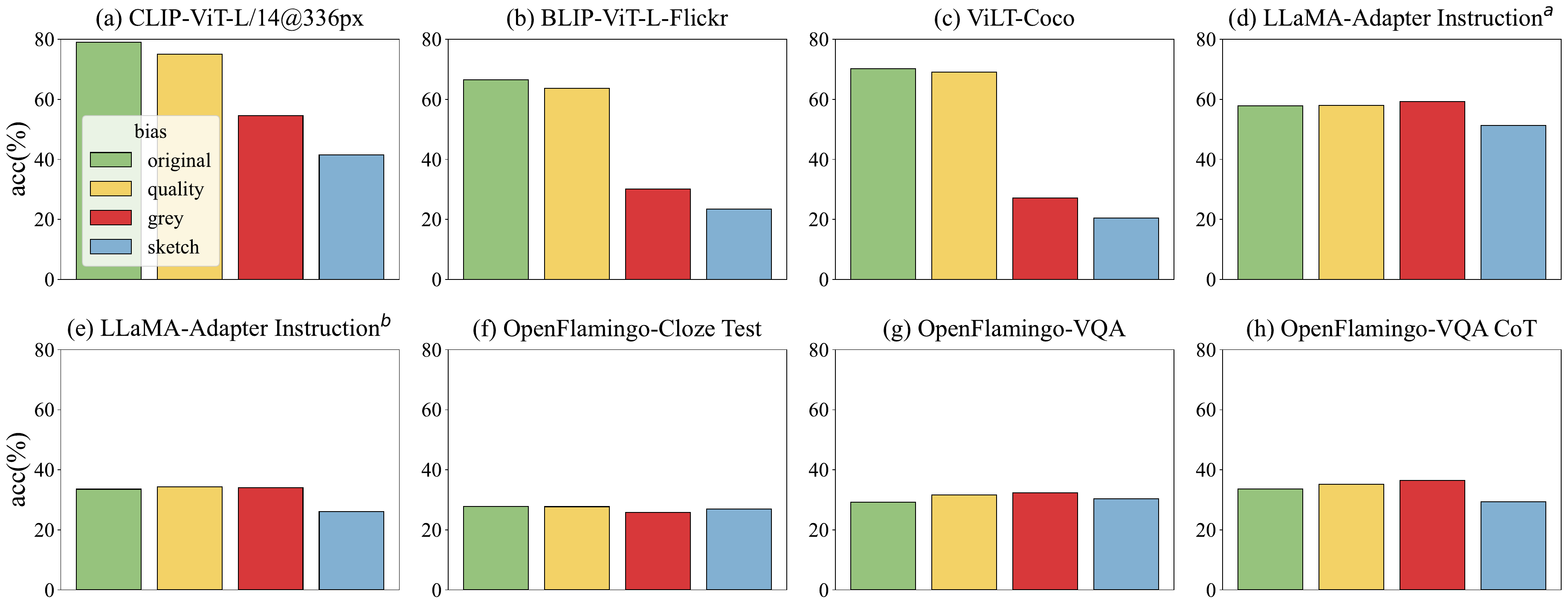}
    \caption{Times}
    \label{fig:bias-time}
\end{subfigure}
\hspace{-0.2cm}
\begin{subfigure}{0.48\textwidth}
    \includegraphics[width=\textwidth]{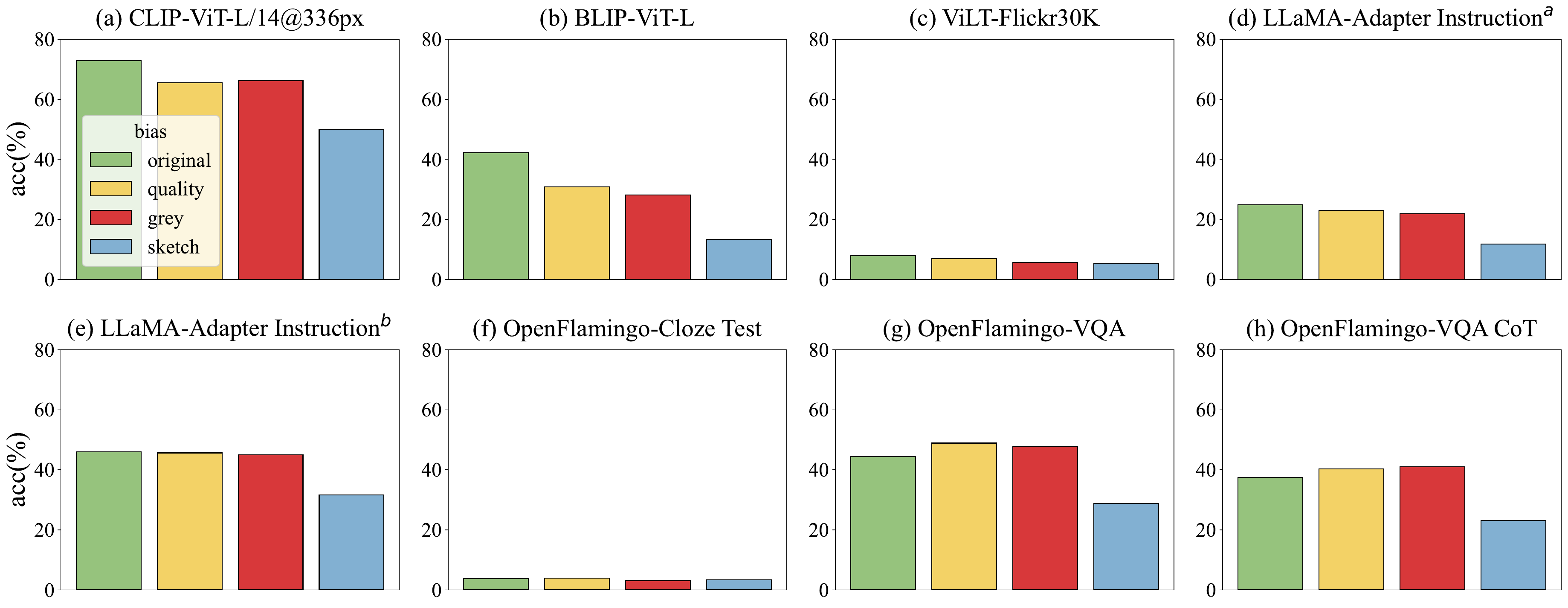}
    \caption{Location(Country)}
    \label{fig:bias-country}
\end{subfigure}
\caption{We investigate how different image biases influence model performance on tasks related to Times and Location (Country). We compare the test set under four settings: original, lower quality, grayscale, and sketch. Our findings reveal that for discriminative VLMs, there is a significant drop in model performance when we change image styles. However, generative VLMs are less affected by these changes. Please zoom in for a better view.}
\label{fig:bias}
\end{figure}

As shown in Fig.~\ref{fig:bias}, lower quality does not really influence the model performance since visual encoders always reshape the input images, and images in sketches are always unrecognizable and significantly deteriorate model performance. However, grayscale images can evidently decrease the performance of discriminative VLMs.

Nevertheless, generative VLMs are almost unaffected by biases. Again, we conducted a failure case study as in Supplementary Material. Generative VLMs are not really grounded by visual cues of images. Answers depend on contexts, such as in-context demonstrations and instructions, and expose the hallucination problem~\cite{ji2023survey}. Therefore, both image details and styles cannot help reasoning.

\vspace{0.2cm}
\section{Conclusion}

In this work, we propose a two-stage probing task \recognition\space and \reasoning\space and introduce a new dataset WikiTiLo, to explore VLMs' times and location reasoning ability 
Based on the findings of \recognition\space and \reasoning\space tasks on the WikiTiLo dataset, we have probed the capabilities of VLMs in times and location reasoning. 
Experiments indicate that discriminative VLMs, such as CLIP, effectively extract context-agnostic visual features for identifying times and locations. However, generative VLMs struggle to reason correctly due to the inability of context-conditioned visual features to retain relevant information or the language models' failure to reason based on visual cues, according to the experiments and case studies. 
This motivates further exploration into how large language models can better utilize visual features in future studies.

\section*{Acknowledgements}
\noindent
This work has been funded by the German Federal Ministry of Education and Research and the Bavarian State Ministry for Science and the Arts. The authors of this work take full responsibility for its content.

\newpage
{
\small
\bibliographystyle{ieee_fullname}
\bibliography{ref}

\begin{thebibliography}{10}\itemsep=-1pt

\bibitem{alain2016understanding}
Guillaume Alain and Yoshua Bengio.
\newblock Understanding intermediate layers using linear classifier probes.
\newblock {\em arXiv preprint arXiv:1610.01644}, 2016.

\bibitem{alayrac2022flamingo}
Jean-Baptiste Alayrac, Jeff Donahue, Pauline Luc, Antoine Miech, Iain Barr, Yana Hasson, Karel Lenc, Arthur Mensch, Katherine Millican, Malcolm Reynolds, et~al.
\newblock Flamingo: a visual language model for few-shot learning.
\newblock {\em Advances in Neural Information Processing Systems}, 35:23716--23736, 2022.

\bibitem{anas_awadalla_2023_7733589}
Anas Awadalla, Irena Gao, Joshua Gardner, Jack Hessel, Yusuf Hanafy, Wanrong Zhu, Kalyani Marathe, Yonatan Bitton, Samir Gadre, Jenia Jitsev, Simon Kornblith, Pang~Wei Koh, Gabriel Ilharco, Mitchell Wortsman, and Ludwig Schmidt.
\newblock Openflamingo, Mar. 2023.

\bibitem{basaj2021explaining}
Dominika Basaj, Witold Oleszkiewicz, Igor Sieradzki, Micha{\l} G{\'o}rszczak, Barbara Rychalska, Tomasz Trzci{\'n}ski, and Bartosz Zieli{\'n}ski.
\newblock Explaining self-supervised image representations with visual probing.
\newblock Freiburg, Germany: International Joint Conferences on Artificial Intelligence, 2021.

\bibitem{brown2020language}
Tom Brown, Benjamin Mann, Nick Ryder, Melanie Subbiah, Jared~D Kaplan, Prafulla Dhariwal, Arvind Neelakantan, Pranav Shyam, Girish Sastry, Amanda Askell, et~al.
\newblock Language models are few-shot learners.
\newblock {\em Advances in neural information processing systems}, 33:1877--1901, 2020.

\bibitem{chen2020uniter}
Yen-Chun Chen, Linjie Li, Licheng Yu, Ahmed El~Kholy, Faisal Ahmed, Zhe Gan, Yu Cheng, and Jingjing Liu.
\newblock Uniter: Universal image-text representation learning.
\newblock In {\em Computer Vision--ECCV 2020: 16th European Conference, Glasgow, UK, August 23--28, 2020, Proceedings, Part XXX}, pages 104--120. Springer, 2020.

\bibitem{lindstrom2021probing}
Adam Dahlgren~Lindstr{\"o}m, Johanna Bj{\"o}rklund, Suna Bensch, and Frank Drewes.
\newblock Probing multimodal embeddings for linguistic properties: the visual-semantic case.
\newblock In {\em Proceedings of the 28th International Conference on Computational Linguistics}, pages 730--744, Barcelona, Spain (Online), Dec. 2020. International Committee on Computational Linguistics.

\bibitem{fu2021violet}
Tsu-Jui Fu, Linjie Li, Zhe Gan, Kevin Lin, William~Yang Wang, Lijuan Wang, and Zicheng Liu.
\newblock Violet: End-to-end video-language transformers with masked visual-token modeling.
\newblock {\em arXiv preprint arXiv:2111.12681}, 2021.

\bibitem{fu2022there}
Xingyu Fu, Ben Zhou, Ishaan Chandratreya, Carl Vondrick, and Dan Roth.
\newblock There’sa time and place for reasoning beyond the image.
\newblock In {\em Proceedings of the 60th Annual Meeting of the Association for Computational Linguistics (Volume 1: Long Papers)}, pages 1138--1149, 2022.

\bibitem{zhang2023llamaadapter}
Peng Gao, Jiaming Han, Renrui Zhang, Ziyi Lin, Shijie Geng, Aojun Zhou, Wei Zhang, Pan Lu, Conghui He, Xiangyu Yue, et~al.
\newblock Llama-adapter v2: Parameter-efficient visual instruction model.
\newblock {\em arXiv preprint arXiv:2304.15010}, 2023.

\bibitem{girdhar2023imagebind}
Rohit Girdhar, Alaaeldin El-Nouby, Zhuang Liu, Mannat Singh, Kalyan~Vasudev Alwala, Armand Joulin, and Ishan Misra.
\newblock Imagebind: One embedding space to bind them all.
\newblock In {\em Proceedings of the IEEE/CVF Conference on Computer Vision and Pattern Recognition}, pages 15180--15190, 2023.

\bibitem{he2016deep}
Kaiming He, Xiangyu Zhang, Shaoqing Ren, and Jian Sun.
\newblock Deep residual learning for image recognition.
\newblock In {\em Proceedings of the IEEE conference on computer vision and pattern recognition}, pages 770--778, 2016.

\bibitem{hewitt2019designing}
John Hewitt and Percy Liang.
\newblock Designing and interpreting probes with control tasks.
\newblock In {\em Conference on Empirical Methods in Natural Language Processing}. Association for Computational Linguistics, 2019.

\bibitem{ji2023survey}
Ziwei Ji, Nayeon Lee, Rita Frieske, Tiezheng Yu, Dan Su, Yan Xu, Etsuko Ishii, Ye~Jin Bang, Andrea Madotto, and Pascale Fung.
\newblock Survey of hallucination in natural language generation.
\newblock {\em ACM Computing Surveys}, 55(12):1--38, 2023.

\bibitem{kim2015predicting}
Hyo~Jin Kim, Enrique Dunn, and Jan-Michael Frahm.
\newblock Predicting good features for image geo-localization using per-bundle vlad.
\newblock In {\em Proceedings of the IEEE International Conference on Computer Vision}, pages 1170--1178, 2015.

\bibitem{kim2021vilt}
Wonjae Kim, Bokyung Son, and Ildoo Kim.
\newblock Vilt: Vision-and-language transformer without convolution or region supervision.
\newblock In {\em International Conference on Machine Learning}, pages 5583--5594. PMLR, 2021.

\bibitem{li2022align}
Dongxu Li, Junnan Li, Hongdong Li, Juan~Carlos Niebles, and Steven~CH Hoi.
\newblock Align and prompt: Video-and-language pre-training with entity prompts.
\newblock In {\em Proceedings of the IEEE/CVF Conference on Computer Vision and Pattern Recognition}, pages 4953--4963, 2022.

\bibitem{li2023blip2}
Junnan Li, Dongxu Li, Silvio Savarese, and Steven Hoi.
\newblock {BLIP-2:} bootstrapping language-image pre-training with frozen image encoders and large language models.
\newblock In {\em ICML}, 2023.

\bibitem{li2022blip}
Junnan Li, Dongxu Li, Caiming Xiong, and Steven Hoi.
\newblock Blip: Bootstrapping language-image pre-training for unified vision-language understanding and generation.
\newblock In {\em International Conference on Machine Learning}, pages 12888--12900. PMLR, 2022.

\bibitem{li2019visualbert}
Liunian~Harold Li, Mark Yatskar, Da Yin, Cho-Jui Hsieh, and Kai-Wei Chang.
\newblock Visualbert: A simple and performant baseline for vision and language.
\newblock {\em arXiv preprint arXiv:1908.03557}, 2019.

\bibitem{lin2013cross}
Tsung-Yi Lin, Serge Belongie, and James Hays.
\newblock Cross-view image geolocalization.
\newblock In {\em Proceedings of the IEEE Conference on Computer Vision and Pattern Recognition}, pages 891--898, 2013.

\bibitem{liu2023visual}
Haotian Liu, Chunyuan Li, Qingyang Wu, and Yong~Jae Lee.
\newblock Visual instruction tuning.
\newblock {\em arXiv preprint arXiv:2304.08485}, 2023.

\bibitem{lu2019vilbert}
Jiasen Lu, Dhruv Batra, Devi Parikh, and Stefan Lee.
\newblock Vilbert: Pretraining task-agnostic visiolinguistic representations for vision-and-language tasks.
\newblock {\em Advances in neural information processing systems}, 32, 2019.

\bibitem{mu2022slip}
Norman Mu, Alexander Kirillov, David Wagner, and Saining Xie.
\newblock Slip: Self-supervision meets language-image pre-training.
\newblock In {\em Computer Vision--ECCV 2022: 17th European Conference, Tel Aviv, Israel, October 23--27, 2022, Proceedings, Part XXVI}, pages 529--544. Springer, 2022.

\bibitem{radford2021learning}
Alec Radford, Jong~Wook Kim, Chris Hallacy, Aditya Ramesh, Gabriel Goh, Sandhini Agarwal, Girish Sastry, Amanda Askell, Pamela Mishkin, Jack Clark, et~al.
\newblock Learning transferable visual models from natural language supervision.
\newblock In {\em International conference on machine learning}, pages 8748--8763. PMLR, 2021.

\bibitem{rosch2023probing}
Philipp~J. R{\"o}sch and Jind{\v{r}}ich Libovick{\'y}.
\newblock Probing the role of positional information in vision-language models.
\newblock In {\em Findings of the Association for Computational Linguistics: NAACL 2022}, pages 1031--1041, Seattle, United States, July 2022. Association for Computational Linguistics.

\bibitem{salem2020learning}
Tawfiq Salem, Scott Workman, and Nathan Jacobs.
\newblock Learning a dynamic map of visual appearance.
\newblock In {\em Proceedings of the IEEE/CVF Conference on Computer Vision and Pattern Recognition}, pages 12435--12444, 2020.

\bibitem{salin2022are}
Emmanuelle Salin, Badreddine Farah, St{\'e}phane Ayache, and Benoit Favre.
\newblock Are vision-language transformers learning multimodal representations? a probing perspective.
\newblock In {\em Proceedings of the AAAI Conference on Artificial Intelligence}, volume~36, pages 11248--11257, 2022.

\bibitem{seo2018cplanet}
Paul~Hongsuck Seo, Tobias Weyand, Jack Sim, and Bohyung Han.
\newblock Cplanet: Enhancing image geolocalization by combinatorial partitioning of maps.
\newblock In {\em Proceedings of the European Conference on Computer Vision (ECCV)}, pages 536--551, 2018.

\bibitem{shi2016does}
Xing Shi, Inkit Padhi, and Kevin Knight.
\newblock Does string-based neural mt learn source syntax?
\newblock In {\em Proceedings of the 2016 conference on empirical methods in natural language processing}, pages 1526--1534, 2016.

\bibitem{singh2022flava}
Amanpreet Singh, Ronghang Hu, Vedanuj Goswami, Guillaume Couairon, Wojciech Galuba, Marcus Rohrbach, and Douwe Kiela.
\newblock Flava: A foundational language and vision alignment model.
\newblock In {\em Proceedings of the IEEE/CVF Conference on Computer Vision and Pattern Recognition}, pages 15638--15650, 2022.

\bibitem{taras2016does}
Vas Taras, Piers Steel, and Bradley~L Kirkman.
\newblock Does country equate with culture? beyond geography in the search for cultural boundaries.
\newblock {\em Management International Review}, 56:455--487, 2016.

\bibitem{touvron2023llama}
Hugo Touvron, Thibaut Lavril, Gautier Izacard, Xavier Martinet, Marie-Anne Lachaux, Timoth{\'e}e Lacroix, Baptiste Rozi{\`e}re, Naman Goyal, Eric Hambro, Faisal Azhar, et~al.
\newblock Llama: Open and efficient foundation language models.
\newblock {\em arXiv preprint arXiv:2302.13971}, 2023.

\bibitem{vo2016localizing}
Nam~N Vo and James Hays.
\newblock Localizing and orienting street views using overhead imagery.
\newblock In {\em Computer Vision--ECCV 2016: 14th European Conference, Amsterdam, The Netherlands, October 11--14, 2016, Proceedings, Part I 14}, pages 494--509. Springer, 2016.

\bibitem{weyand2016planet}
Tobias Weyand, Ilya Kostrikov, and James Philbin.
\newblock Planet-photo geolocation with convolutional neural networks.
\newblock In {\em Computer Vision--ECCV 2016: 14th European Conference, Amsterdam, The Netherlands, October 11-14, 2016, Proceedings, Part VIII 14}, pages 37--55. Springer, 2016.

\bibitem{yang2022unified}
Jianwei Yang, Chunyuan Li, Pengchuan Zhang, Bin Xiao, Ce Liu, Lu Yuan, and Jianfeng Gao.
\newblock Unified contrastive learning in image-text-label space.
\newblock In {\em Proceedings of the IEEE/CVF Conference on Computer Vision and Pattern Recognition}, pages 19163--19173, 2022.

\bibitem{yao2021filip}
Lewei Yao, Runhui Huang, Lu Hou, Guansong Lu, Minzhe Niu, Hang Xu, Xiaodan Liang, Zhenguo Li, Xin Jiang, and Chunjing Xu.
\newblock Filip: fine-grained interactive language-image pre-training.
\newblock {\em arXiv preprint arXiv:2111.07783}, 2021.

\bibitem{yu2022coca}
Jiahui Yu, Zirui Wang, Vijay Vasudevan, Legg Yeung, Mojtaba Seyedhosseini, and Yonghui Wu.
\newblock Coca: Contrastive captioners are image-text foundation models.
\newblock {\em arXiv preprint arXiv:2205.01917}, 2022.

\bibitem{zhang2023llama}
Renrui Zhang, Jiaming Han, Aojun Zhou, Xiangfei Hu, Shilin Yan, Pan Lu, Hongsheng Li, Peng Gao, and Yu Qiao.
\newblock Llama-adapter: Efficient fine-tuning of language models with zero-init attention.
\newblock {\em arXiv preprint arXiv:2303.16199}, 2023.

\end{thebibliography}
}

\end{document}


\title{Can Vision-Language Models be a Good Guesser? \\
Exploring VLMs for Times and Location Reasoning \\
Supplementary Materials\\}

\def\institutey{LMU Munich}
\def\institutew{Munich Center for Machine Learning}
\author{
\textbf{Gengyuan Zhang \textsuperscript{1,2} \quad Yurui Zhang \textsuperscript{3} \quad Kerui Zhang \textsuperscript{1} \quad Volker Tresp \textsuperscript{1,2}}  \\
\textsuperscript{1} LMU Munich, Munich, Germany\\
\textsuperscript{2} Munich Center for Machine Learning, Munich, Germany \\
\textsuperscript{3} Technical University of Munich \\
\tt\small zhang@dbs.ifi.lmu.de 
}

\maketitle

\appendix

\section{Dataset WikiTiLo}
\subsection{List of countries in WikiTiLo}
The countries included in WikiTiLo and their regions are listed in Tab.~\ref{tab:country area}. These countries are almost evenly distributed in 7 regions defined by their cultural and geographical affinity with reference of UNESCO\footnote{https://population.un.org/wpp/DefinitionOfRegions/} and sorted alphabetically.

\begin{table}[h]
\centering
  \resizebox{0.45\textwidth}{!}{
  \begin{tabular}{|l|l||l|l|}
    \hline \makebox[1.5cm]{$\mathbf{Country}$} & \makebox[1cm]{$\mathbf{Region}$} & \makebox[1.5cm]{$\mathbf{Country}$} & \makebox[1cm]{$\mathbf{Region}$}\\
    \hline\hline Afghanistan &  Middle East & Argentina & Latin America\\
    \hline Australia &  NA, EU and OC & Brazil &  Latin America \\
    \hline Bangladesh &  Southern Asia & China & Eastern Asia \\
    \hline Germany &  NA, EU and OC & India & Southern Asia   \\
    \hline Indonesia &  South-Eastern Asia & Iran & Middle East \\
    \hline Japan &  Eastern Asia  & Kazakhstan & Central Asia\\
    \hline Kenya &  Sub-Saharan Africa  & Kyrgyzstan & Central Asia\\
    \hline Malaysia & South-Eastern Asia &  Mexico & Latin America\\
    \hline Nigeria & Sub-Saharan Africa & North Korea & Eastern Asia \\
    \hline Pakistan & Middle East & Rwanda & Sub-Saharan Africa  \\
    \hline Saudi Arabia & Middle East &  South Africa & Sub-Saharan Africa\\
    \hline South Korea & Eastern Asia & Sri Lanka & Southern Asia \\
    \hline Tajikistan & Central Asia & Thailand & South-Eastern Asia  \\
    \hline Turkmenistan & Central Asia & United States & NA, EU and OC  \\
    \hline Uzbekistan & Central Aisa & Vietnam & South-Eastern Asia  \\
    \hline
  \end{tabular}}
  \caption{Countries with corresponding regions (NA, EU, and OC is the abbreviation of North America, Europa, and Oceania).}
  \label{tab:country area}
\end{table}

\subsection{Data curation}
In order to guarantee that WikiTiLo comprises images that are characteristics of socio-cultural visual hints, we conduct a manual image curation based on image visual cues on raw images in Wikimedia Commons, as in Fig.~\ref{fig:analyse1}. We try to ensure that the space identity and time period of each
image can be distinguished from the architectural patterns, costume styles, language types, movement postures, photo colors, and quality, or other fine-grained features.

\begin{figure}
\centering
\includegraphics[width=0.5\textwidth]{result_data/Image Reasoning.pdf}  
\caption{An example of manual image curating to determine its time and location. By doing this, the images of WikiTiLo are grounded by multiple scene text, faces, object segments from the image, and colors and resolution of the image. }
\label{fig:analyse1}
\end{figure}

\subsection{Data distribution}
The dataset distribution in location and times can be found in Fig.~\ref{fig:distribution}.

\begin{figure}[h]
\hspace{0.7cm}
    \begin{subfigure}[]{0.36\textwidth}
    \centering
      \includegraphics[trim=0 0 0 0, clip, width=0.83\textwidth]{CLIP_Result_PDF/Space Distribution.pdf}
      \caption{Location Distribution}
    \end{subfigure}

\hspace{0.7cm}
    \begin{subfigure}[]{0.36\textwidth}
    \centering
\includegraphics[trim=0 1.2cm 0.2cm 1cm, clip,  width=\textwidth]{CLIP_Result_PDF/Time Distribution.pdf}
      \caption{Times Distribution}
    \end{subfigure}
    \caption{The WikiTiLo dataset exhibits a diverse distribution of times and locations. In terms of time, the images range from 1827 to the post-2000 era. Regarding location, we selected 30 countries whose images met our filtering criteria, representing eight regions. Due to the development of Internet media, images taken after 2000 constitute a significant portion of the dataset. Conversely, images taken before 1900 only account for 10\% of the dataset, primarily due to limited data availability and poor quality.}
    \label{fig:distribution}
  \end{figure}

\section{Visual encoders of discriminative VLMs}
We compare all the visual encoders of discriminative Vision Language Models and Vision Models we used for references in the paper in the dimension of the dataset, visual encoder, and textual encoder in Tab.~\ref{tab:models}.
\begin{table}[h]
    \centering
    \resizebox{0.5\textwidth}{!}{
    \begin{tabular}{l|cccc}
    \toprule
         Model & Visual Encoder & Text Encoder & Dataset  & Multimodal\\
         \hline
         ResNet-50 & ResNet & -  & ImageNet~\cite{russakovsky2015imagenet} & \ding{56} \\
         ViLT & Patch Emb. & BERT & MSCOCO~\cite{lin2014microsoft}, GCC~\cite{sharma2018conceptual} & \ding{52} \\
        CLIP-ViT & ViT & Transf. & Online data & \ding{52} \\
        CLIP-RN & ResNet & Transf.  & Online data & \ding{52} \\
        BLIP & ViT & BERT  & CapFilt~\cite{li2022blip},14M\cite{li2022blip}, 129M\cite{li2022blip} & \ding{52} \\
        
    \bottomrule
    \end{tabular} }
    \caption{Comparison of disminitative VLMs. Variants of different datasets and encoders will be denoted by suffixes.}
    \label{tab:models}
\end{table}

\section{Impact of shot numbers}
\begin{figure}[H]
\centering
    \hspace{-0.3cm}
\includegraphics[width=0.46\textwidth]{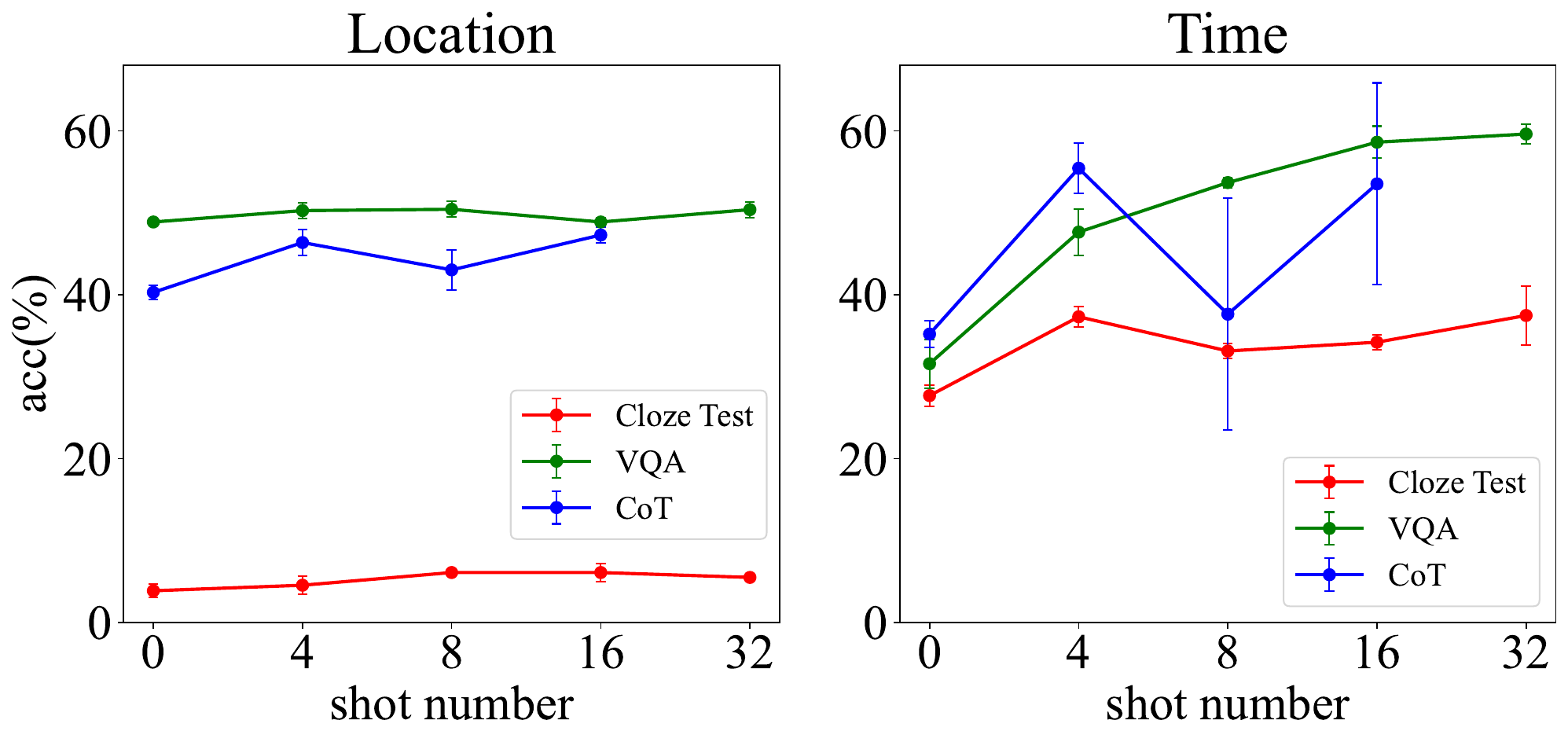}
\caption{Impact of different shot numbers for OpenFlamingo. More in-context shots do not substantially increase the performance on \reasonloc, but achieve a higher accuracy on \reasontimes.}
\label{fig:shot}
\end{figure}

We studied the impact of shot number for OpenFlamingo as in Fig.~\ref{fig:shot}.
Especially for \reasontimes, we find the output prediction is more unstable when having more in-context shots and deteriorates the performance.

\section{Visualization}
We show the visualization of the transportation plan of word patch alignment on times classification as Fig.7 in the main body. For Times-relevant questions, the attended patches seem less specific. Generally,
visual tokens in the background instead of foreground objects have seemingly dominant contributions.

\begin{figure*}[h]
\vspace{-0.2cm}
\begin{subfigure}[b]{0.24\textwidth}
\vspace{-0.1cm}
\includegraphics[width=\textwidth]{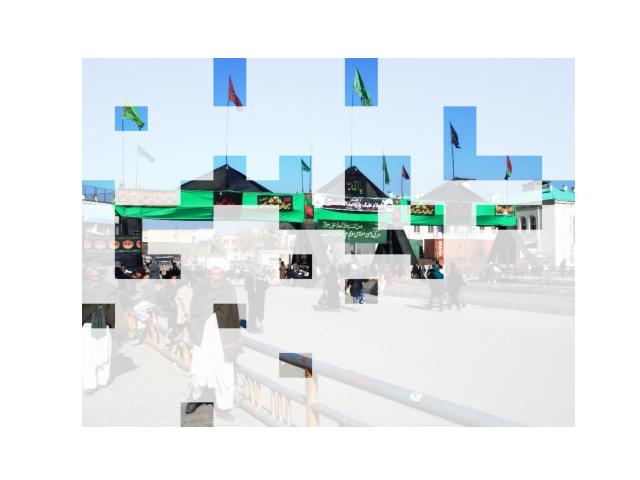}
\vspace{-0.8cm}
\caption{}
\end{subfigure}
\begin{subfigure}[b]{0.24\textwidth}
\vspace{-0.1cm}
\includegraphics[width=\textwidth]{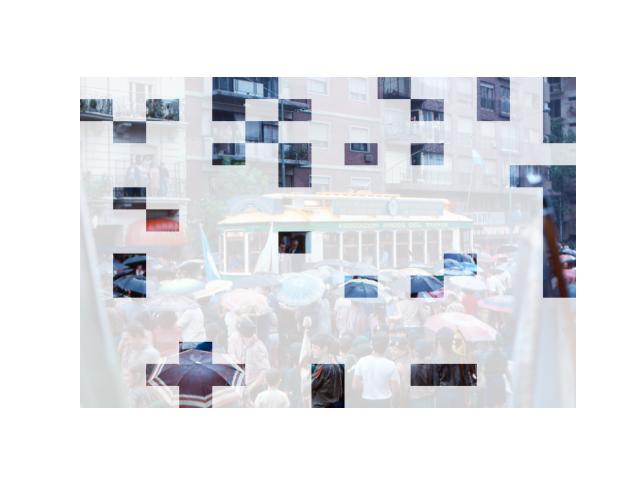}
\vspace{-0.8cm}
\caption{}
\end{subfigure}
\begin{subfigure}[b]{0.24\textwidth}
\vspace{-0.1cm}
\includegraphics[width=\textwidth]{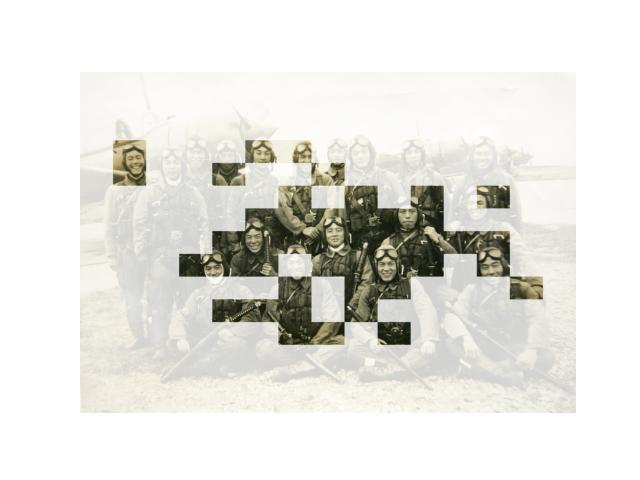}
\vspace{-0.8cm}
\caption{}
\end{subfigure}
\begin{subfigure}[b]{0.24\textwidth}
\vspace{-0.1cm}
\includegraphics[width=\textwidth]{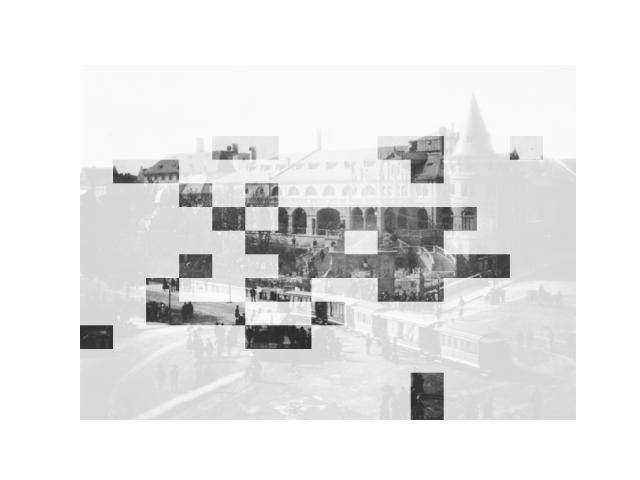}
\vspace{-0.8cm}
\caption{}
\end{subfigure}

\begin{subfigure}[b]{0.24\textwidth}
\vspace{-0.1cm}
\includegraphics[width=\textwidth]{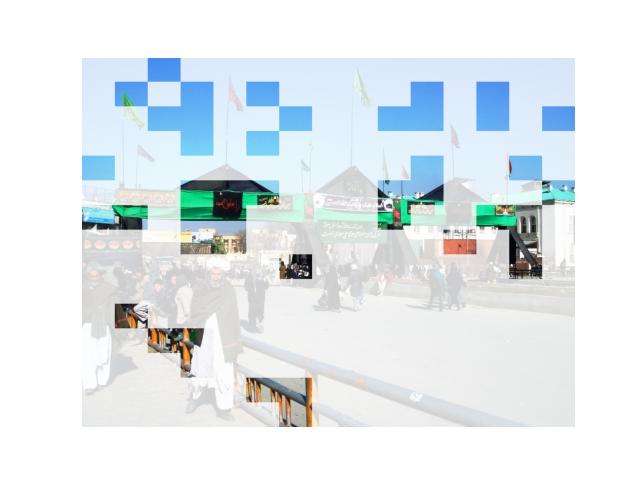}
\vspace{-0.8cm}
\caption{}
\end{subfigure}
\begin{subfigure}[b]{0.24\textwidth}
\vspace{-0.1cm}
\includegraphics[width=\textwidth]{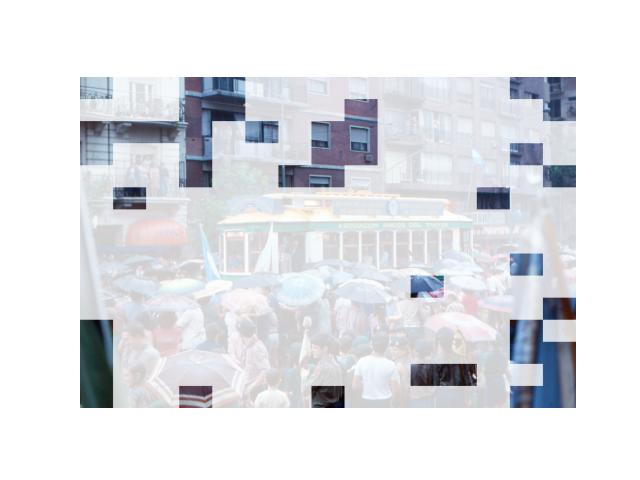}
\vspace{-0.8cm}
\caption{}
\end{subfigure}
\begin{subfigure}[b]{0.24\textwidth}
\vspace{-0.1cm}
\includegraphics[width=\textwidth]{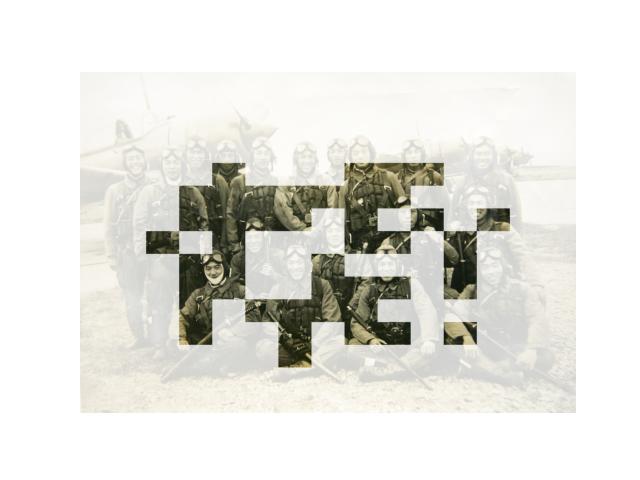}
\vspace{-0.8cm}
\caption{}
\end{subfigure}
\begin{subfigure}[b]{0.24\textwidth}
\vspace{-0.1cm}
\includegraphics[width=\textwidth]{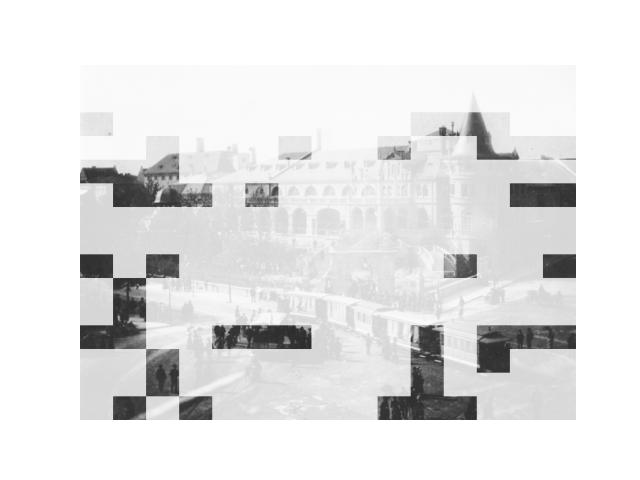}
\vspace{-0.8cm}
\caption{}
\end{subfigure}

\begin{subfigure}[b]{0.24\textwidth}
\vspace{-0.1cm}
\includegraphics[width=\textwidth]{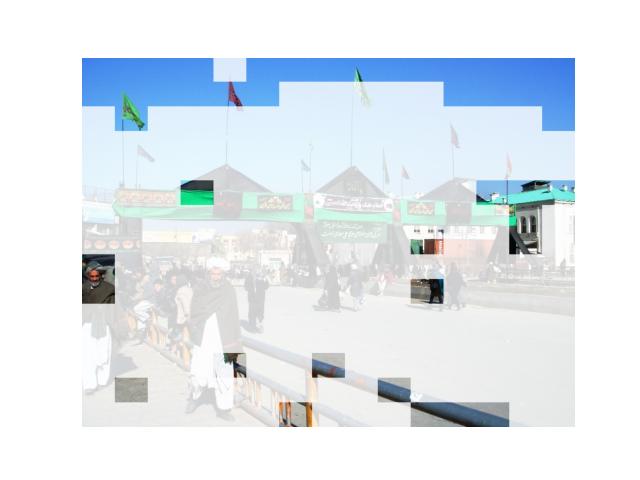}
\vspace{-0.8cm}
\caption{}
\end{subfigure}
\begin{subfigure}[b]{0.24\textwidth}
\vspace{-0.1cm}
\includegraphics[width=\textwidth]{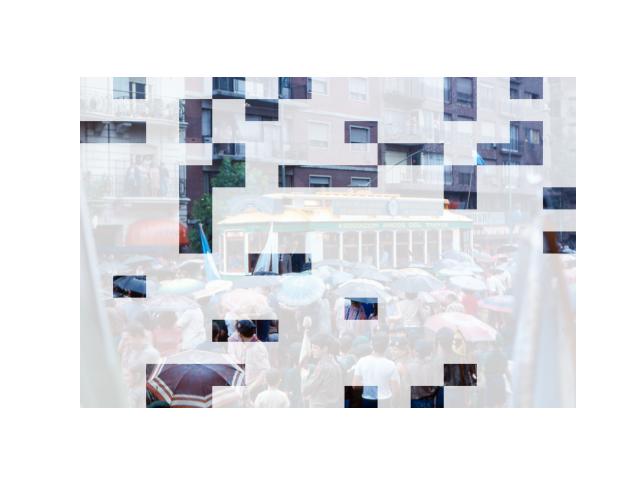}
\vspace{-0.8cm}
\caption{}
\end{subfigure}
\begin{subfigure}[b]{0.24\textwidth}
\vspace{-0.1cm}
\includegraphics[width=\textwidth]{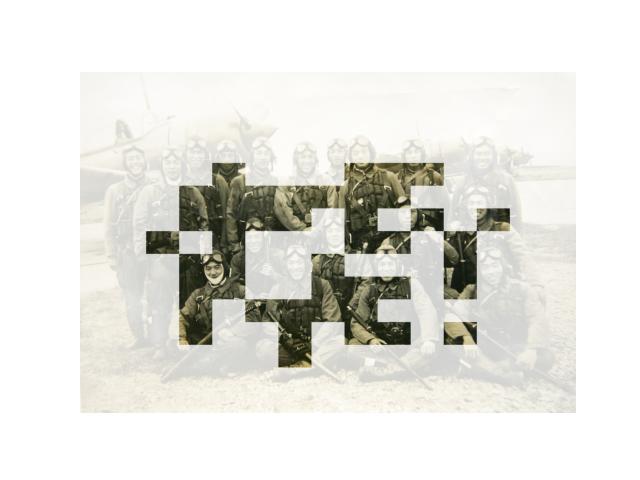}
\vspace{-0.8cm}
\caption{}
\end{subfigure}
\begin{subfigure}[b]{0.24\textwidth}
\vspace{-0.1cm}
\includegraphics[width=\textwidth]{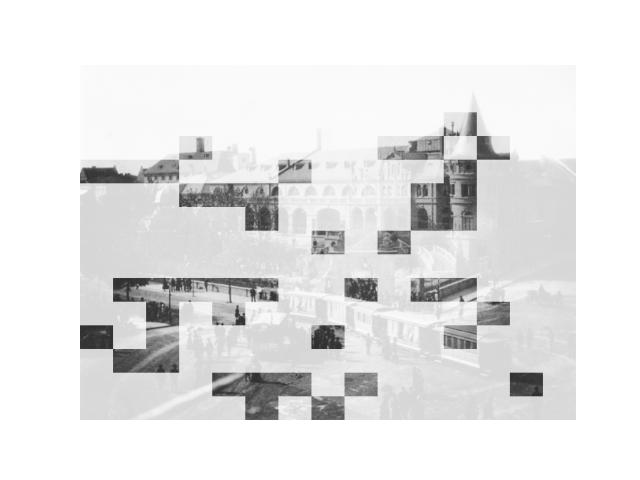}
\vspace{-0.8cm}
\caption{}
\end{subfigure}
\caption{Visualization of transportation plan of word patch alignment on times classification. Best viewed zoomed in. Rows from top to bottom: ViLT, CLIP, and BLIP. Columns from left to right: Afghanistan(Middle East) in 2000, Argentina(Latin America) in 1980, Japan(Eastern Asia) in 1940, and Germany(Europe) in 1880.}
\label{fig:vis_times}
\end{figure*}

\section{Prompts used for generative VLMs on  \textsc{reasoning}$_\textsc{Times}$}
We also list all the prompts used for OpenFlamingo and LLaMA-Adapter V2 used for \textsc{reasoning}$_\textsc{Times}$ in Fig.~\ref{fig:instruction} as in the paper for references. 

\begin{figure}
\hspace{-0.5cm}
\centering
\begin{subfigure}[b]{0.45\textwidth}
\scalebox{0.67}{
\fbox{
\begin{minipage}{35em}
\textbf{OpenFlamingo Cloze Test} \\
\texttt{<image>}
\textit{Output}: This is a historical photo taken in the 19th Century. \texttt{<\textbar endofchunk\textbar>}\\
\textbf{Open Flamingo VQA}\\
We divide time into 4 eras. These 4 eras are in the 19th Century, between 1900 and 1950, between 1950 and 2000, in the 21st Century.\\
\texttt{<image>}
\textit{Question}: When was this photo taken? \\ 
\textit{Short answer}: in the 21st Century
\texttt{<\textbar endofchunk\textbar>}\\
\textbf{OpenFlamingo VQA - CoT} \\
We divide time into 4 eras. These 4 eras are in the 19th Century, between 1900 and 1950, between 1950 and 2000, in the 21st Century.\\
\texttt{<image>}
\textit{Question}: When was this photo taken? \\
\textit{Answer}: Because the people in this photograph are dressed in attire typical of the Qing Dynasty in China. Therefore, it can be inferred that this photograph was taken during the Qing Dynasty,this photo was taken in the 19th Century.\texttt{<\textbar endofchunk\textbar>}
\end{minipage}}}
\caption{}
\end{subfigure}

\hspace{-0.5cm}
\begin{subfigure}[h]{0.45\textwidth}
\centering
\scalebox{0.67}{
\fbox{
\centering
\vspace{-2cm}
\begin{minipage}{35em}
\textbf{LLaMA-Adapter V2 Instruction$^a$} \\\textit{Instruction}: This photograph was taken during one of the following 4 periods. We divide these 4 periods as in the 19th Century, between 1900 and 1950, between 1950 and 2000, in the 21st Century. In which period was this photo taken?\\
\textbf{LLaMA-Adapter V2 Instruction$^b$} \\\textit{Instruction}: In which period was this photo taken?
\end{minipage}}}
\label{fig:instruction-times}
\caption{}
\end{subfigure}
\label{fig:prompt-times}
\caption{We list respectively the prompt templates we used in OpenFlamingo for each protocol for \textsc{Reasoning} for times in (a), instructions for LLaMA-Adapter V2 for times in (b).}
\label{fig:instruction}
\end{figure}

\begin{figure}
\hspace{-0.5cm}
\centering
\begin{subfigure}[b]{0.45\textwidth}
\scalebox{0.67}{
\fbox{
\begin{minipage}{35em}
\textbf{OpenFlamingo Cloze Test} \\
\texttt{<image>}
\textit{Output}: This is a local photo taken in area Latin America.
\texttt{<\textbar endofchunk\textbar>}\\
\textbf{Open Flamingo VQA}\\
The photograph was taken in one of the following eight areas. These eight areas are "Central Asia," "Southern Asia," "Latin America," "Northern America, Europe and Oceania," "Middle East," "Eastern Asia," "South-Eastern Asia," "Sub-Saharan Africa." \\
\texttt{<image>}
\textit{Question}: In which area was this photograph taken? \\ 
\textit{Short answer}: Southern Asia
\texttt{<\textbar endofchunk\textbar>}\\
\textbf{OpenFlamingo VQA - CoT} \\
The photograph was taken in one of the following eight areas. These 8 areas are "Central Asia," "Southern Asia," "Latin America," "Northern America, Europe and Oceania," "Middle East," "Eastern Asia," "South-Eastern Asia," "Sub-Saharan Africa." \\
\texttt{<image>}
\textit{Question}: In which area was this photograph taken? \\
\textit{Answer}: Because in the photo, there is a man wearing a turban, and the photo includes a mosque, this photo was taken in the Middle East. \texttt{<\textbar endofchunk\textbar>}
\end{minipage}}}
\caption{}
\end{subfigure}

\hspace{-0.5cm}
\begin{subfigure}[h]{0.45\textwidth}
\centering
\scalebox{0.67}{
\fbox{
\centering
\vspace{-2cm}
\begin{minipage}{35em}
\textbf{LLaMA-Adapter V2 Instruction$^a$} \\\textit{Instruction}: The photograph was taken in one of the following eight regions. These eight regions are "Latin America," "Northern America, Europe and Oceania," ... "Eastern Asia," "South-Eastern Asia," "And Sub-Saharan Africa." \\
\textbf{LLaMA-Adapter V2 Instruction$^b$} \\\textit{Instruction}: In which geopolitical region was this photo taken?
\end{minipage}}}
\label{fig:instruction-region}
\caption{}
\end{subfigure}
\label{fig:prompt-region}

\caption{We list the prompt templates we used in location reasoning for OpenFlamingo in (a) and for LLaMA-Adapter V2 in (b).}
\end{figure}





\section{Rationale examples for Chain-of-Thought}
\begin{figure}[]
    \centering
    \includegraphics[trim=0 0 0 0, clip, width=0.5\textwidth]{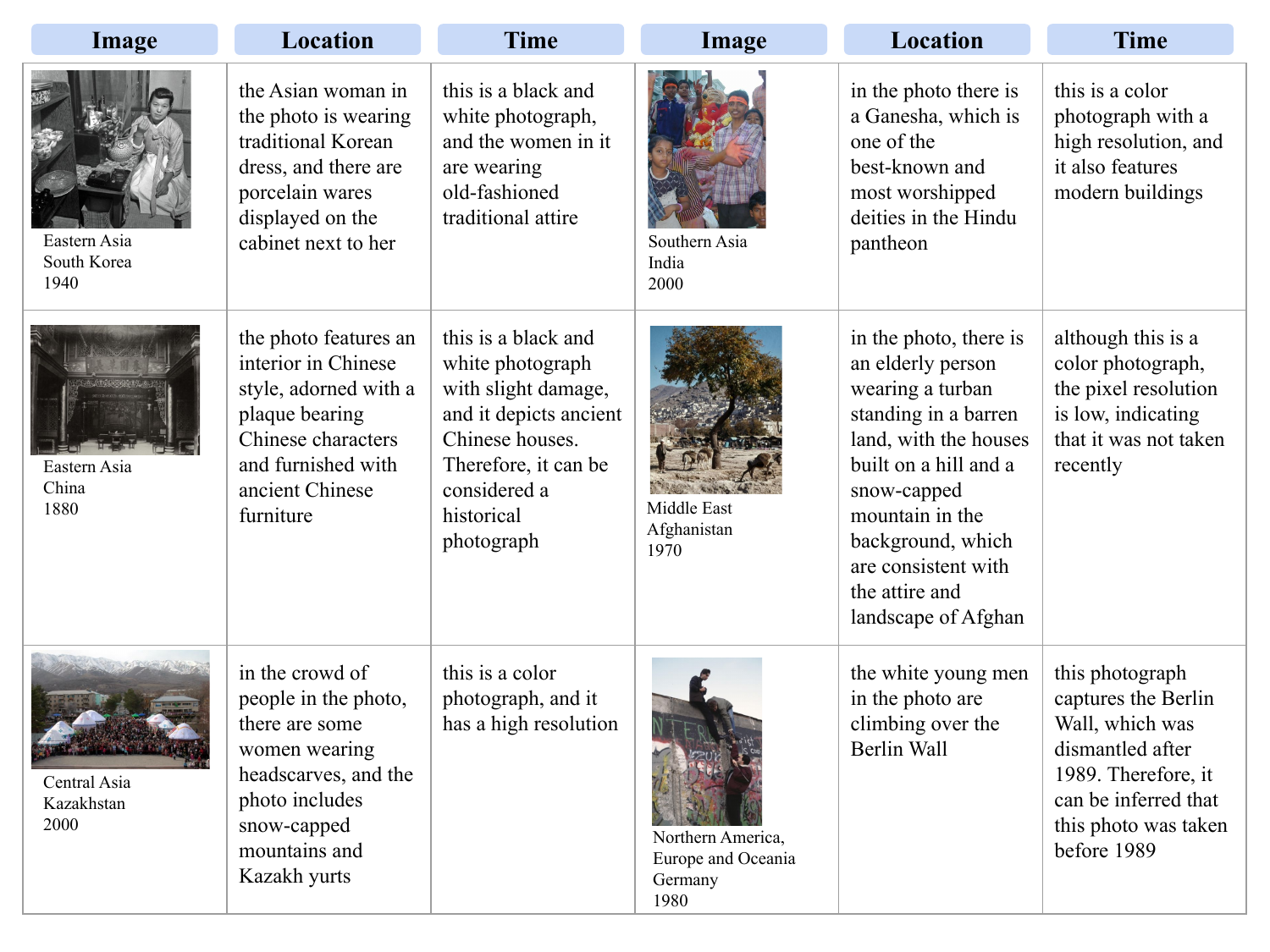}
    \caption{Rationale examples we used as few shot samples in protocol OpenFlamingo VQA with CoT.}
    \label{fig:cot-examples}
  \end{figure}
We annotate a subset of images with rationale in \textsc{Reasoning} tasks for OpenFlamingo Chain-of-Thought. Here, we showcase some examples of images and the rationale associated. We attempt to include visual details that are relevant for reasoning about times and locations for humans.

\begin{figure}[]
    \centering
    \includegraphics[width=0.48\textwidth]{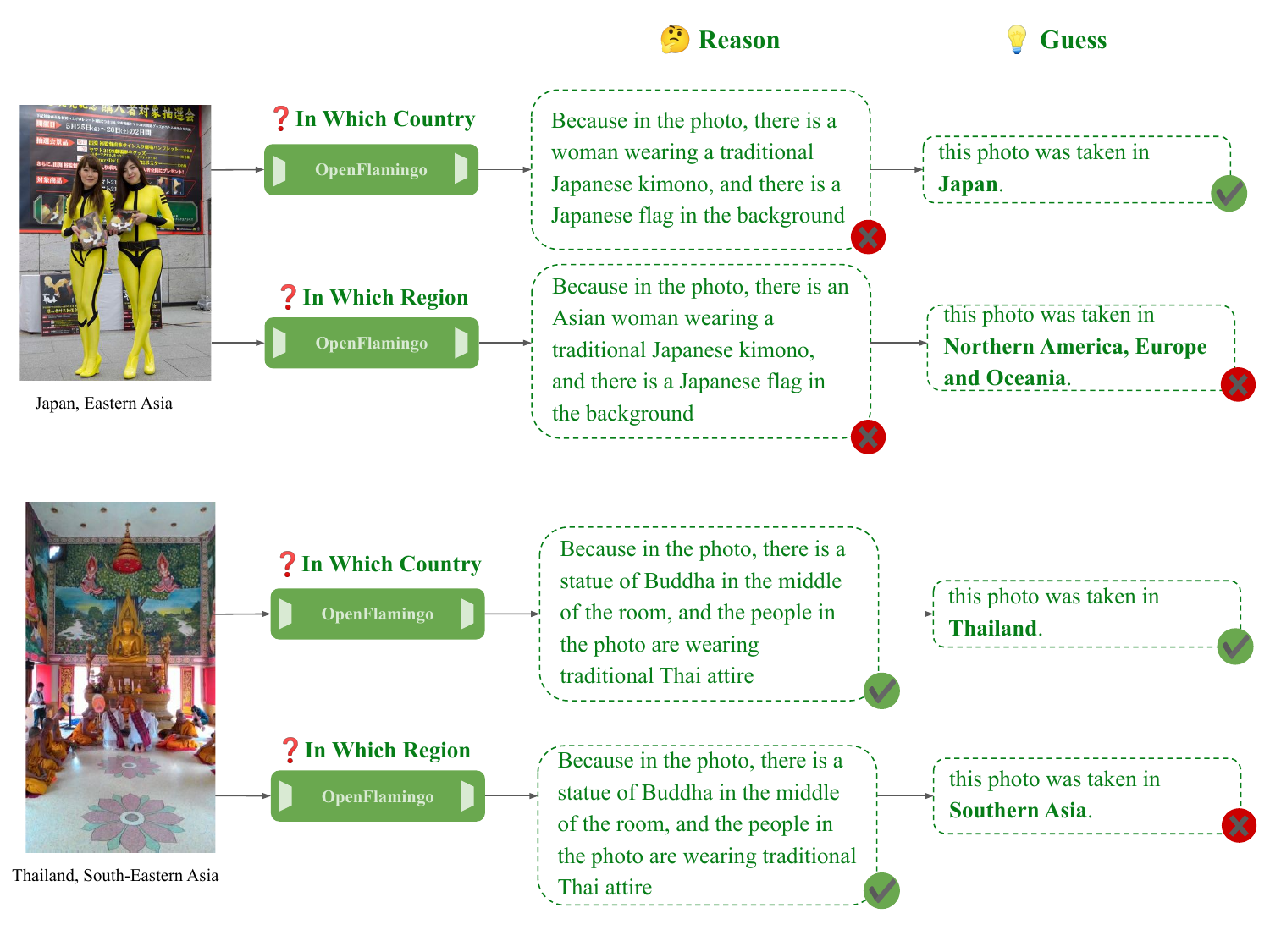}
  \caption{In some cases, the model can correctly predict the country level but fails in the region reasoning. We speculate that the visual cues could not be mapped exactly to the region definitions in our experiments.}
   \label{fig:region-country}
\end{figure}

\section{Case study}
\subsection{Failure on reasoning regions}
We observe that generative VLMs perform worse in reasoning regions than reasoning countries, which is against intuition. We conduct a qualitative case study on the failure cases as in Fig.~\ref{fig:region-country}. It is shown that generative VLMs actually cannot really ground the reasoning process, especially in two-step reasoning. Even if the model gives correct reasoning that implies the country, it still fails to correlate to the corresponding countries. This again shows us the performance of models contained by the language models.

\subsection{Dataset bias}
We compare the model prediction of original images and transfer the images into three common image styles: low quality, grayscale, and sketch. 
We select several example photos and show how generative VLMs fail in these cases. We find predictions of generative VLMs are not really grounded by visual cues of images. Answers depend on contexts, such as in-context demonstrations and instructions, and expose the hallucination problem~\cite{ji2023survey}. Therefore, image details and style biases cannot help or influence the model reason.

\begin{figure}[]
    \centering
    \includegraphics[width=0.5\textwidth]{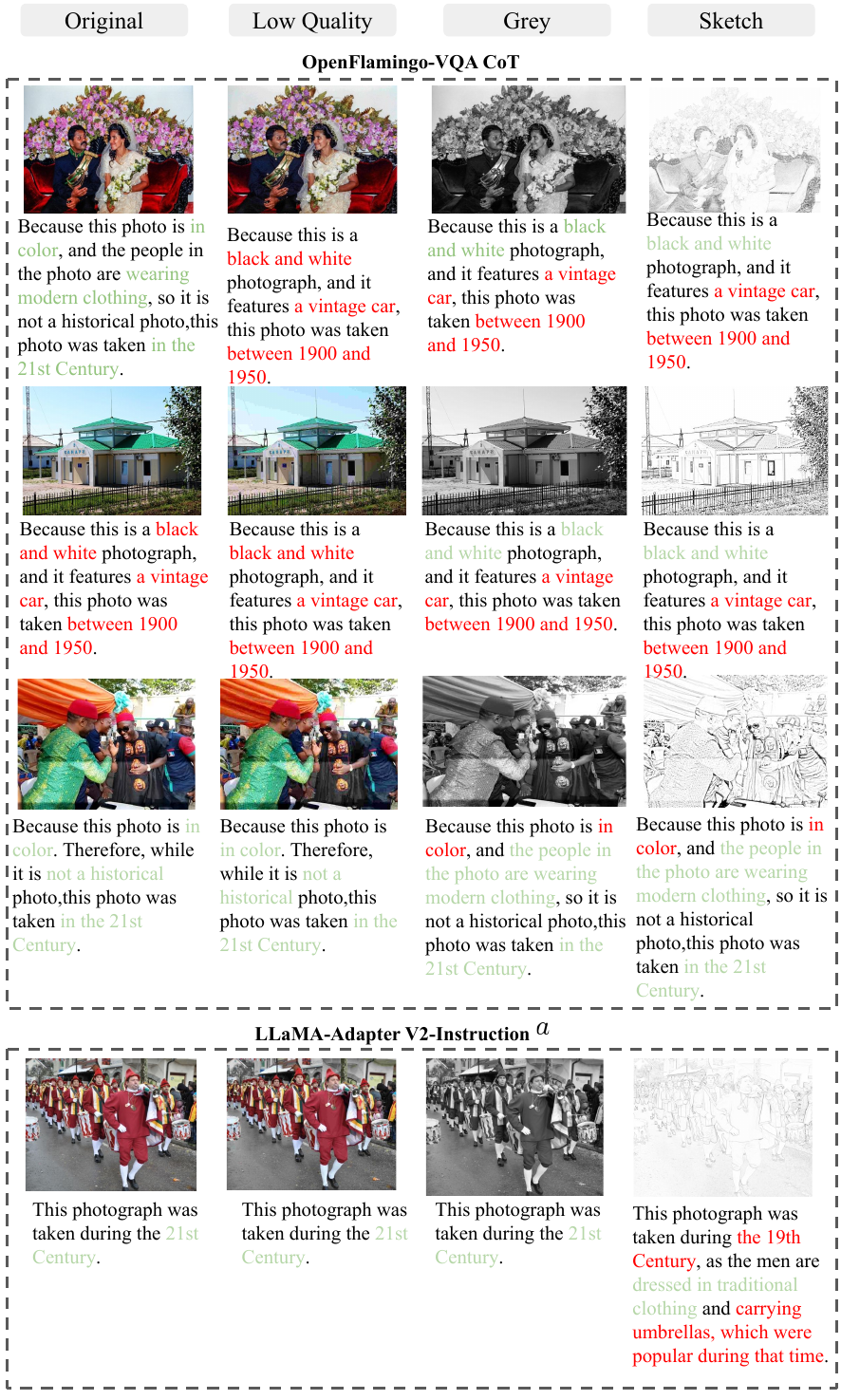}
  \caption{Case study of generative VLMs under different image biases. We compare the model prediction of original images and transfer the images into three common image styles that prevail in the dataset: low quality, grayscale, and sketch. The factual generation is marked in \textcolor{green}{green}, and the wrong generation is in \textcolor{red}{red}.}
   \label{fig:bias-case}
\end{figure}

\newpage
{\small
\bibliographystyle{ieee_fullname}
\bibliography{ref}
}
\newpage